\documentclass[english]{article}
\usepackage[utf8]{inputenc}
\usepackage[T1]{fontenc}
\usepackage{babel}
\usepackage{amsmath}
\usepackage{graphicx}
\usepackage{fancyhdr}
\usepackage{multirow}
\usepackage{hyperref}
\usepackage{longtable}
\usepackage{siunitx}
\usepackage{eurosym}
\usepackage{amssymb}
\usepackage{arydshln}
\usepackage{comment}
\usepackage{titlesec}
\usepackage{algorithm}
\usepackage{algorithmic}
\usepackage[a4paper, total={6in, 9in}]{geometry}
\hypersetup{
    colorlinks=true,
    linkcolor=blue,
    filecolor=magenta,      
    urlcolor=cyan,
    pdfpagemode=FullScreen,
}
\usepackage{caption}
\usepackage{subcaption}
\usepackage{svg}
\svgsetup{inkscapepath=svgsubdir}

\begin{document}

\title{Unified machine learning tasks and datasets for enhancing renewable energy}
\author{Arsam Aryandoust\textsuperscript{1†*}, Thomas Rigoni\textsuperscript{2†}, Francesco di Stefano\textsuperscript{2†}, \\ 
Anthony Patt\textsuperscript{2}}

\maketitle

\textsuperscript{1}MIT, Laboratory for Information \& Decision Systems. *corresponding author: arsam@mit.edu \\

\textsuperscript{2}ETH Zurich, Climate Policy Lab. \textsuperscript{†}Co-first authors. \\

\begin{abstract}
    Multi-tasking machine learning (ML) models exhibit prediction abilities in domains with little to no training data available (few-shot and zero-shot learning). Over-parameterized ML models are further capable of zero-loss training and near-optimal generalization performance. An open research question is, how these novel paradigms contribute to solving tasks related to enhancing the renewable energy transition and mitigating climate change. A collection of unified ML tasks and datasets from this domain can largely facilitate the development and empirical testing of such models, but is currently missing. Here, we introduce the \emph{ETT-17} (\emph{E}nergy \emph{T}ransition \emph{T}asks-17), a collection of 17 datasets from six different application domains related to enhancing renewable energy, including out-of-distribution validation and testing data. We unify all tasks and datasets, such that they can be solved using a single multi-tasking ML model. We further analyse the dimensions of each dataset; investigate what they require for designing over-parameterized models; introduce a set of dataset scores that describe important properties of each task and dataset; and provide performance benchmarks. 
\end{abstract}

\newpage
\section*{Introduction}

The electrification of our energy consumption and the supply of this electricity from renewable sources like wind and solar is one of the most effective ways for opposing global warming and mitigating climate change \cite{Shukla2022}. This energy transition requires significant technological and societal changes \cite{Patt2015}, as well as tools for realising these changes within the short period of time that is left for reducing global greenhouse gas (GHG) emissions and preventing irreversible losses to our lives and ecosystems \cite{Arias2023}. Artificial intelligence (AI) and its sub-field of machine learning (ML) are increasingly recognized to be powerful computational tools with an immense potential for contributing to the technologies, policies, and social processes required for a successful transition to renewable energy, as well as for climate change mitigation and adaptation in a much broader application space \cite{Rolnick2022}.

Recent advancements in AI and ML show that large-scale multi-tasking models often outperform their smaller and more specialized counterparts \cite{Brown2020, Sanh2021, Reed2022}. These models excel at leveraging transfer learning across related and unrelated domains, acquiring powerful prediction abilities that enable them to generalize surprisingly well in yet unprecedented tasks where data for training a model is limited or not available at all, something that is known as \emph{few-shot} and \emph{zero-shot learning}. Initially developed for natural language processing, these models are known as \emph{large language models} (LLMs), with ChatGPT being one of their most powerful and prominent representatives. Building upon the fundamental elements of LLMs, a variety of \emph{foundation models} \cite{Lu2021, Bommasani2022} and \emph{generalist agents} \cite{Janner2021, Chen2021, Reid2022, Zheng2022, Furuta2022, Reed2022} have been developed to address a wide range of multi-tasking problems in domains beyond natural language processing.

Another important paradigm in AI are highly over-parameterized ML models. These are models with largely more parameters than data points they are trained on, which are often capable of zero-loss training, a phenomenon called \emph{interpolation}. Over-parameterized ML models are further found to achieve near-optimal generalization accuracy \cite{Belkin2019}, even under high levels of noisy data \cite{Zhang2017}, and are able to adopt desired properties like associative memory \cite{Radhakrishnan2020} and high robustness \cite{Bubeck2023}.

Current research in AI, however, has a significant disparity: While large attention is given to designing powerful multi-tasking ML models like Gato \cite{Reed2022} and ChatGPT \cite{Brown2020} and over-parameterized ML models for mainstream tasks from natural language processing, robotics, and computer vision, only little attention is paid to designing such models for tasks from climate change-related domains. Consequently, the advancement of AI models in climate change-related domains is lagging behind and has mainly remained a by-product of endeavors in other application domains. This discrepancy is already resulting in the lack of suitable tools for addressing climate change challenges with AI, which often involve themes like spatio-temporal Earth science data and physics-informed ML \cite{Tu2022}. 

A major reason for this discrepancy is that we currently do not understand the properties that are prevalent in climate change-related tasks and datasets. If an ML model does not perform well on a specific task, we do not understand what attributes of a task and its associated data have led to this outcome. A comprehensive collection and analysis of tasks and datasets related to climate change will therefore greatly facilitate the design of new ML solutions in climate change domains, and potentially lead to important advancements in research on more general ML methods \cite{Donti2022, Aryandoust2023}.

A number of studies contribute to this direction. Rolnick et al. create a comprehensive overview of ML applications for tackling climate change, which also includes a comprehensive set of applications crucial for advancing renewable energy \cite{Rolnick2022}. Mosavi et al., and later Donti \& Kolter, review numerous applications of ML for facilitating the development and operation of sustainable energy \cite{Mosavi2019, Donti2021}. Nguyen et al. developed the first foundation model for ML tasks in weather and climate modelling \cite{Nguyen2023}, making a more imminent contribution to multi-tasking AI models in a climate-change related domain. Koh et al. introduce a collection of 10 datasets, of which some are related to tackling climate change, and which represent common real-world data distribution shifts, aiming to address their under-representation in currently popular ML tasks; they demonstrate substantial performance differences of default ML models on \emph{in-distribution} and \emph{out-of-distribution} (ood) data, highlighting the need for designing more robust ML models and for evaluating these on ood data \cite{Koh2021}.

\begin{figure}[ht]
    \centering
    \includegraphics[height = 11cm]{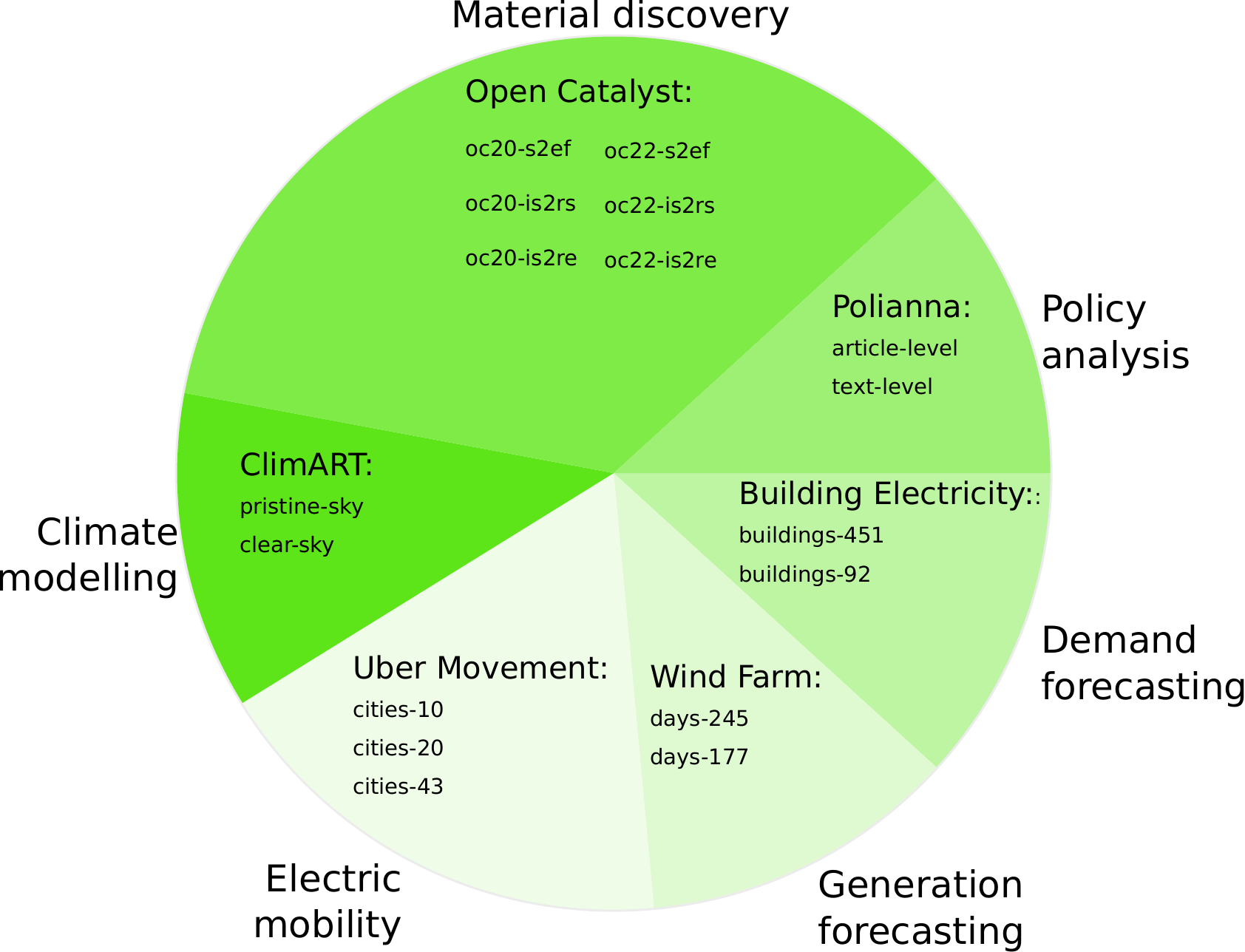}
    \caption{Overview of application domains, ML tasks, sub-task datasets of the ETT-17 collection.}
    \label{figure:tasks_and_datasets}
\end{figure}

In this work, we present 17 datasets that span across six parent application domains with a focus on tackling climate change by enhancing the renewable energy transition. Figure \ref{figure:tasks_and_datasets} provides an overview of the application domains to which each collected task and sub-task dataset belongs. These include applications for demand forecasting based on satellite imagery, generation forecasting of an entire wind farm, traffic prediction for electrifying mobility, increasing resolution of weather and climate models, discovering catalysts for hydrogen production and combustion cells, and analysing climate and energy policy articles. Our objective is to provide engineers with a valuable resource for designing and testing advanced AI models in this domain. To facilitate the development of new models, mainly those involving multi-tasking and transfer learning scenarios that process diverse data modalities and configurations, we have devised a unified data format based on spatial and temporal components of datasets. This standardized format simplifies the use of our datasets and the design of large models by bringing these onto a common ground. The collection of datasets we prepare is publicly available (\url{https://dataverse.harvard.edu/dataverse/EnergyTransitionTasks}), and can be loaded in different formats using a Python software package (\url{https://github.com/Selber-AI/selberai}). In the following sections, we first introduce each task and its corresponding sub-task dataset, employing representative names to facilitate clarity throughout our work. Next, we present our unified data representation and describe the analysis we conduct for characterising each dataset. Then, we present and discuss the outcome of our analysis, and describe our methodology in extensive detail.

\subsubsection*{Collected tasks and datasets: Energy Transition Tasks 17 (ETT-17)}

The tasks and datasets we collect are important for the global energy transition in different ways. Here, we briefly introduce each task and their respective sub-task datasets as presented in Figure \ref{figure:tasks_and_datasets}.

\textit{Building Electricity}: Given the aerial image of a building and the meteorological conditions in the region of that building for a past time window of 24h, we want to predict the electric load profile of single buildings for a future time window of 24h \cite{Aryandoust2022}. 
Such demand forecasting tasks are crucial for planning the dispatch of renewable power systems on short timescales \cite{Aryandoust2023_DR}. 
We distinguish two sub-task datsets, one containing 92 buildings which we refer to as the \textit{buildings-92} dataset, and the other containing 451 buildings which we refer to as the \textit{builidngs-451} dataset.

\textit{Wind Farm}: Given two weeks of measured values of each turbine on a wind farm and the relative geographic position of all turbines in the farm to each other, we want to predict the active power generation of the entire farm for two days into the future \cite{Zhou2022}. 
Solving this task and knowing the accurate generation of wind farms provides a great advantage for integrating high shares of renewable wind energy into our current power systems at a higher pace.
We distinguish two sub-task datasets, one containing data from all turbines for 245 days which we refer to as the \textit{days-245} dataset, and the other containing data for 177 days which we refer to as the \textit{days-177} dataset.

\textit{Uber Movement}: Given a pair of origin-destination zones of a city for a certain time stamp, we want to predict the average travel time that a car would need for this trip \cite{Aryandoust2019}. 
Predicting these travel times and solving this task is useful for planning the distribution and sizing of charging stations in renewable power systems when electrifying mobility and utilizing storage capacities for vehicle-to-grid applications.
We distinguish three sub-tasks containing data from 10, 20 and 43 cities which we respectively refer to as the \textit{cities-10}, \textit{cities-20} and \textit{cities-43} datasets.

\textit{ClimART}: Given the concentration of gases and aerosols at different layers of the atmosphere, we want to predict the atmospheric radiative transfer (ART) between each layer \cite{Cachay2021}. 
This allows us to develop surrogate AI models for well-understood physics-based Earth system models, and expand their resolution in time and space through faster computation. These improvements allow us to have more accurate planning and dispatch of wind, solar and other renewable generation units.
We distinguish two sub-task datasets, that is, the \textit{pristine-sky} dataset where only the concentration of gases is considered and the \textit{clear-sky} dataset, where also the concentration of aerosols is considered. 

\textit{Open Catalyst}: Given the initial structure of arbitrary catalyst-adsorbate pairs for hydrogen electrolysis and fuel cells, we want to predict the relaxed structure and energy \cite{Zitnick2020, Chanussot2021, Tran2022}.
This allows us to develop surrogate AI models for density functional theory (DFT) simulations, which allows us to accelerate the discovery of novel cost-effective catalysts with desired properties. 
We distinguish six sub-task datasets, of which half belong to the \emph{Open Catalyst 2020} project and the other half belong to the \emph{Open Catalyst 2022} project. In the first set of sub-tasks which we refer to as \emph{oc20-s2ef} and \emph{oc22-s2ef}, the per atom forces and overall system energy has to be predicted from a 3-dimensional atomic structure of a adsorbate-catalyst system. In the second set of sub-tasks which we refer to as \emph{oc20-is2rs} and \emph{oc22-is2rs}, the relaxed structure has to be predicted from an initial structure of a catalyst-adsorbate system. In the third set of sub-tasks which we refer to as \emph{oc20-is2re} and \emph{oc22-is2re}, only the relaxed structure's energy has to be predicted from an initial structure of a catalyst-adsorbate system. 

\textit{Polianna}: Given the article of a policy directive or regulation, we want to annotate text spans in the article with a set of (i) policy instrument types, (ii) policy design characteristics, and (iii) technology and application specifications \cite{Sewerin2023}. 
Enhancing solutions to this task allows us to develop new semi-automated analytical tools for assessing the effectiveness of climate and energy policy designs, and move away from resource-intensive hand-coding by policy analysts in this domain.
We distinguish two different sub-task datasets which we refer to as the \textit{article-level} task and the \textit{text-level} task. In the \textit{article-level} task, we want to estimate the frequency of annotations for the overall text of an article, regardless of their precise positions within the text. In the \textit{text-level} task, we instead want to identify precise text spans within an article that belong to a particular annotation.

\subsubsection*{Unified data representation and analysis}

Unifying tasks and datasets allows us to process multiple data modalities using a single representation and a single, or only a few, multi-tasking ML model(s). It further facilitates the comparison and analysis of dataset characteristics across different tasks and domains. Here, we briefly introduce our unified data representation, describe how we analyse requirements for designing over-parameterized models, what properties we quantify using dataset scores, and how we benchmark prediction results. 

We standardize data components in (virtual) time and space to formulate a unified data format that is able to bring all tasks and datasets onto a common ground. We assign each data point, consisting of a feature and a label vector, to a canonically defined point in (virtual) time and space. We distinguish data points by three separate parts: one that varies only across time; one part that varies only across space; and one part that varies across both time and space. Supplementary Information 1 further compares this representation with alternative existing unified data formats.

The over-parameterized regime is reached when the number of ML model parameters exceeds the number of training data points times the dimension of data point labels \cite{Belkin2019}, a value that is called the \emph{interpolation threshold} ($ipt$). This is the point at which theory suggests that we reach maximum over-fitting to our data and minimum generalization performance for the number of ML model parameters we can explicitly choose, known as the \emph{interpolation peak} \cite{Belkin2019}. Increasing the number of ML model parameters beyond the $ipt$ allows us to experience a \emph{double descent} of \emph{true risk}, where generalization performance starts to increase again and often exceeds performance of ML models from the under-parameterized regime \cite{Belkin2019}. Within the region of over-parameterized ML models, theory suggests that a second threshold for \emph{robust} or \emph{smooth} functional relationships learnt by an ML model can be expected when the number of parameters starts to exceed the $ipt$ times the dimension of data point features, something known as the \emph{universal law of robustness} \cite{Bubeck2023}. We call this second value the \emph{smooth function threshold} ($sft$). In order to compute the $ipt$ and $sft$ for each sub-task dataset, we compute the number of data points $n$, the ratios by which we split these into training/validation/testing $s_{tr}$/$s_{va}$/$s_{te}$, as well as the dimension of features $D_x$ and labels $D_y$.

A problem with designing trustworthy AI and ML models for real-world applications is given when these are not evaluated sufficiently on data from regions that are not seen during training (\emph{Epistemic uncertainty}), or when very similar inputs have drastically different outputs (\emph{Aleatoric uncertainty}). The performance of AI and ML models for a particular application further depends on another set of data characteristics, such as imbalanced representations that do not reflect the real world (\emph{selection bias}) or insufficiently evaluated subgroups and edge cases (\emph{evaluation bias}). These are challenges that often occur in data that is not \emph{identically and independently distributed} (\emph{non-iid}), such as in data distributions that change across time and space. To quantify these properties and get a better sense of the challenges involved for solving each task that we collect, we introduce four metrics for scoring our datasets with regard to these. All scores are designed to have a value between 0 and 1: 

The first metric we introduce is what we call the sample imbalance score (\emph{SImb-score}), which tells us how large data imbalances and sample biases are that a model sees during training, validation and testing. This allows us to quantify an estimate of selection biases in our datasets. The second metric we introduce is the spatio-temporal out-of-distribution score (\emph{STood-score}), which quantifies the distribution shift of data points for our unified data format across time and space between the training set and the validation and testing sets. This allows us to quantify an estimate of the Epistemic uncertainty associated with each dataset. The third metric we introduce is the input-output score (\emph{IO-score}), which captures the average sensitivity of single labels with respect to changes in single features. This allows us to quantify an estimate of Aleatoric uncertainty in data. The fourth metric we introduce is the \emph{Outlier-score}, which captures the presence of subgroups and edge cases in data. It allows us to quantify an estimate of evaluation biases given in our datasets.

In order to design AI and ML models for solving the collected tasks in real-world applications, we must evaluate performance beyond simple accuracy. We therefore evaluate also energy consumption and computational time for the benchmark results that we compute. Most of our tasks consist of (multi-label) regression problems, while some are classification problems. We create benchmark results for both types of tasks using a \emph{Random Forest (RF)}, which represents a simple to implement, yet highly expressive prediction model. This allows us to gain insight into the complexity of each task with regard to both computation and prediction. For the regression problems, we evaluate prediction performance using the \emph{R2-score}. The \emph{R2-score} can have values between minus infinity and one, where bigger values indicate a better performance. An \emph{R2-score} of 0 represents predictions equal to the mean of all labels. For the classification problems, we evaluate prediction performance using the accuracy measured by the share of correctly classified samples. Our benchmarks, however, are not directly applicable to variable length data representations. Therefore, we omit the creation of benchmarks for these cases. More specialized models like \emph{Recurrent Neural Networks} or \emph{Transformers} could be employed to solve these tasks. However, using these here greatly hinders the idea of creating a simple baseline or benchmark.

\section*{Results}

\subsubsection*{Requirements for over-parameterized ML models}

Table \ref{table:overview} contains the number of data points $n$, the share we split data into training $s_{tr}$, validation $s_{va}$ and testing $s_{te}$, the dimension of features $D_x$ and labels $D_y$, as well as the resulting number of ML model parameters required for reaching the interpolation threshold $ipt$ and the smooth function threshold $sft$ for every ETT-17 task that we collect.

We can observe that the number of data points available for solving each task varies significantly for different domains. The difference between the task with the smallest $n$ (Polianna) and the largest $n$ (Uber Movement) is about seven orders of magnitude. In applications with variable length features and labels, the label dimensions $D_y$ can be both smaller (Polianna) and larger (Polianna and Open Catalyst) than in all other tasks with fixed-length data. As a consequence, the differences for reaching $ipt$ are also large, reaching from a value as small as 0.004M (Polianna) to a value as large as 10B (Uber Movement). There are further differences between the $ipt$ of different sub-tasks for the same application domain. The difference between the smallest and largest $ipt$ in the same application domain can be as small as a factor of 1.4 (= 350M/253M, Wind Farm) and as big as a factor of 440,000 (= 8.8B/0.02M, Open Catalyst). Similar variations can be observed for values of $sft$, because the dimensions of features slightly correlate with the dimensions of labels across all tasks. Values of $sft$ vary between 9.8M (Polianna) and 33T (Open Catalyst). The difference between the smallest and largest $sft$ in the same application domain can be as small as a factor of 1.3 (= 1.6T/1.2T, Wind Farm) and as big as a factor of 402,000 (= 33T/82M, Open Catalyst).

\begin{table} [ht]
\scalebox{0.9}{
\begin{tabular}{|l|l|r|r|c|c|r|r|}
     Application & Dataset & $n$ & $s_{tr}/s_{va}/s_{te}$ & $D_x$ & $D_y$ & $ipt$ & $sft$ \\ \hline
     Building Electricity & buildings-92 & 3,206,016 & 56/09/35 \% & 521 &  96 & 172M & 90B \\
     & buildings-451& 15,716,448 &  56/09/35 \% & 521 & 96 & 845M & 440B \\  
     Wind Farm & days-245 & 3,517,359 & 35/13/52 \% & 4,610 & 288 & 355M & 1.6T \\
     & days-177 &  2,583,966 & 34/13/53 \% & 4,610 & 288 & 253M & 1.2T \\ 
     Uber Movement & cities-10 & 1,037,785,339 & 23/17/60 \% & 11 & 4 & 955M & 10.5B \\
     & cities-20 & 3,266,646,911 & 35/14/51 \% & 11 & 4 & 4.6B & 50.3B \\
     & cities-43 & 7,351,030,412 & 34/14/52 \% & 11 & 4 & 10B & 110B \\  
     ClimART & pristine-sky & 11,452,416 & 51/12/37 \% & 970 & 298 & 1.7B & 1.7T \\
     & clear-sky & 11,485,184 & 51/11/38 \% & 2,487 & 298 & 1.7B & 4.3T \\  
     Open Catalyst & oc20-s2ef & 137,933,475 & 23/08/69 \% & 245-7,775 & 22-576 & 8.8B & 33T \\
     & oc20-is2rs &  560,182 & 27/10/63 \% & 245-7,775 & 21-575 & 42M & 158B \\
     & oc20-isre & 560,182 & 27/10/63 \% & 245-7,775 & 1 & 0.15M & 569M \\
     & oc22-s2ef & 9,070,689 & 36/07/57 \% & 595-7,980 & 52-685 & 1B & 3.8T \\
     & oc22-is2rs & 51,294 & 42/19/39 \% & 595-8,225 & 51-705 & 7M & 26.9B \\
     & oc22-isre & 51,294 & 42/19/39 \% & 595-8,225 & 1 & 0.02M & 82M \\   
     Polianna & article-level & 430 & 22/16/62 \%& 17-4,963 & 42 & 0.004M & 9.8M \\ 
     & text-level & 430 & 22/16/62 \% & 17-4,963 & 3-2,208 & 0.1M & 258M \\   
\end{tabular}}
\caption{\label{table:overview} An analysis of the structure and dimension of each collected prediction task and dataset. $n$ := number of data points in dataset. $s_{tr}/s_{va}/s_{te}$ := shares of $n$ split for training, validation and testing. $D_x/D_y$ := dimension of features/labels. $ipt$ := number of model parameters necessary for reaching interpolation threshold for DL models. $sft$ := number of model parameters necessary for reaching smoothness or robustness threshold of functions in the over-parameterized regime. Units M stand for million, B for billion, and T for trillion.}
\end{table}

\subsubsection*{Dataset property scores}

\begin{figure}[ht]
    \centering
    \includegraphics[height = 20cm]{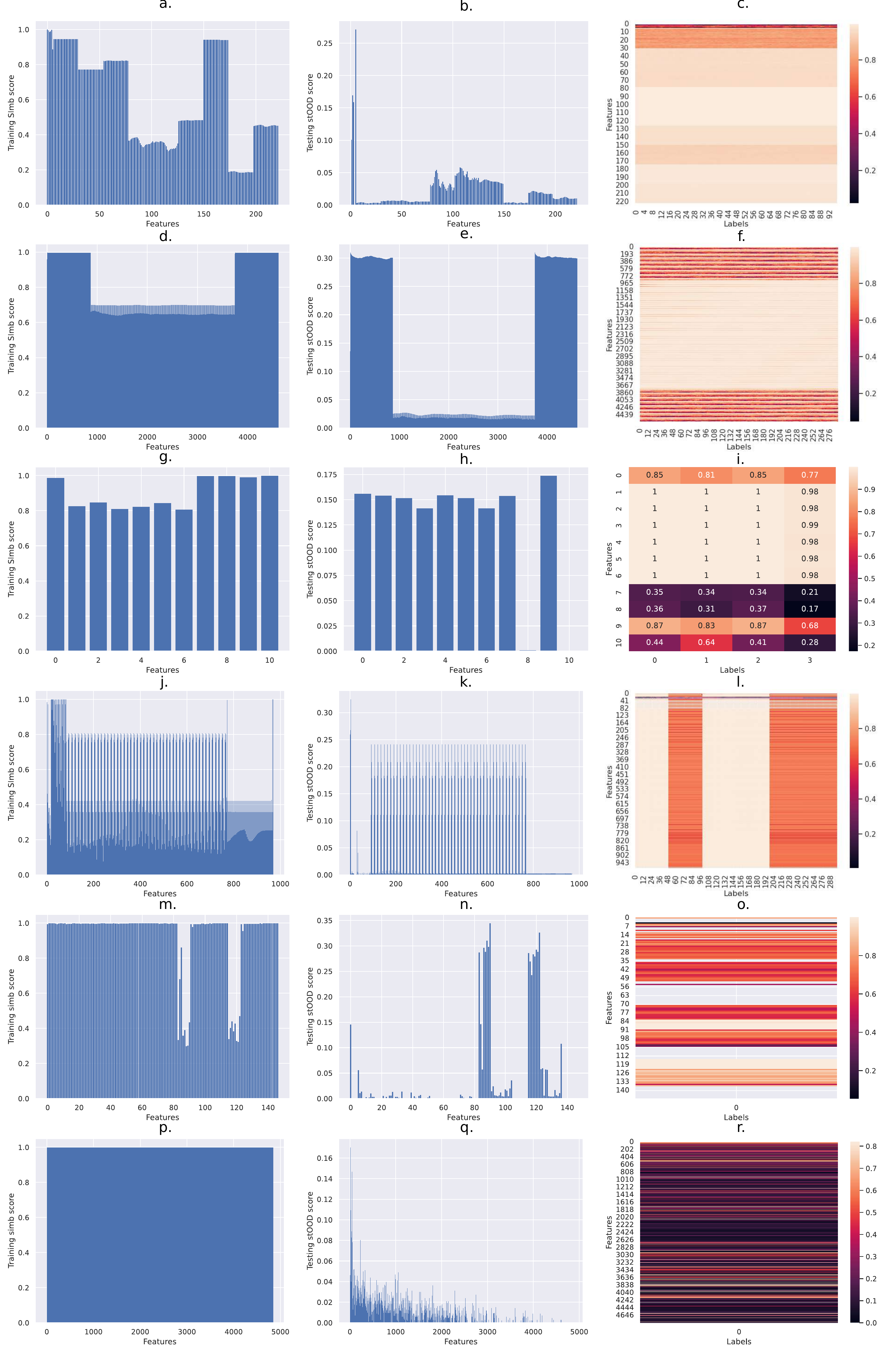}
    \caption{Dataset scores for all tasks. Each row contains results for exemplar sub-task dataset of a task, in the same order as presented in the results tables. Left column contains Training SImb-score, middle column contains Testing STood-score, right column contains IO-score.}
    \label{figure:results_fig}
\end{figure}

Table \ref{table:analysis_scores} contains the resulting dataset scores for each dataset we analyse. These are computed as averages over sub-scores.  Figure \ref{figure:results_fig} contains a plot of these scores representative for each task, which visualize these sub-scores in bar plots and heat-maps to gain a better understanding of the individual variability present in our final scores. Supplementary Information 2 further provides score plots for each analysed dataset in figures \ref{figure:suppinf_be}-\ref{figure:suppinf_pa}. 

We observe large differences of dataset properties between prediction tasks of different application domains as shown in figure \ref{figure:results_fig}, but consistent similarities between dataset properties of sub-tasks from the same application domains as shown in figures \ref{figure:suppinf_be}-\ref{figure:suppinf_pa}. Sample imbalances (SImb-score) can be similar for features and labels (Wind Farm, Uber Movement, Open Catalyst, Polianna), but also largely different between features and labels (Building Electricity, ClimART). Similarly, data distribution shifts across space-time can be very similar in features and labels (Building Electricity, Wind Farm, ClimART), but again largely different between these (Uber Movement, Open Catalyst, Polianna). The sensitivity of labels in response to changes in features (IO-score) spans from relatively low scores of 0.286 (Polianna) to a score as high as 0.941 (Building Electricity), but is commonly around a score of 0.67-0.9. The presence of outliers (Outlier-score) is very consistent across training, validation and testing sets. However, the scores span from relatively low values of 0.045 (Uber Movement) to values as high as 0.801 (Polianna), with common scores being balanced between 0.5-0.6.

\begin{table} [ht]
    \scalebox{0.67}{
        \begin{tabular}{|l|l|cc|cc|c|c|}
        Application & dataset & \begin{tabular}[c]{@{}c@{}}SImb-score\\ features \\ train/val/test\end{tabular} & \begin{tabular}[c]{@{}c@{}}SImb-score\\ labels\\ train/val/test\end{tabular} & \begin{tabular}[c]{@{}c@{}}STood-score\\ features \\ val/test\end{tabular} & \begin{tabular}[c]{@{}c@{}}STood-score\\ labels\\ val/test\end{tabular} & IO-score & \begin{tabular}[c]{@{}c@{}}Outlier-score\\ train/val/test/overall\end{tabular} \\ \hline
        Building Electricity & buildings-92 & 0.132/0.140/0.134 & 0.280/0.269/0.262 & 0.022/0.017 & 0.075/0.071 & 0.913 & 0.52/0.529/0.529/0.515 \\
         & buildings-451 & 0.127/0.131/0.130 & 0.275/0.295/0.293 & 0.009/0.008 & 0.063/0.062 & 0.941 & 0.509/0.499/0.499/0.505 \\
        Wind Farm & days-177 & 0.115/0.120/0.115 & 0.110/0.117/0.112 & 0.010/0.009 & 0.009/0.005 & 0.903 & 0.548/0.549/0.549/0.549 \\
         & days-245 & 0.114/0.117/0.114 & 0.140/0.144/0.140 & 0.007/0.004 & 0.006/0.003 & 0.897 & 0.539/0.536/0.537/0.537 \\
        Uber Movement & cities-10 & 0.330/0.314/0.301 & 0.309/0.310/0.310 & 0.287/0.269 & 0.008/0.007 & 0.733 & 0.114/0.135/0.127/0.045 \\
         & cities-20 & 0.371/0.301/0.299 & 0.314/0.312/0.312 & 0.207/0.202 & 0.008/0.008 & 0.765 & 0.160/0.110/0.082/0.147 \\
         & cities-43 & 0.304/0.251/0.248 & 0.312/0.312/0.312 & 0.200/0.198 & 0.004/0.004 & 0.788 & 0.183/0.204/0.200/0.218 \\
        ClimART & pristine & 0.123/0.124/0.124 & 0.271/0.272/0.272 & 0.002/0.001 & 0.002/0.001 & 0.855 & 0.514/0.516/0.513/0.515 \\
         & clear-sky & 0.115/0.117/0.116 & 0.269/0.270/0.270 & 0.002/0.001 & 0.002/0.001 & 0.854 & 0.593/0.590/0.591/0.592 \\
        Open Catalyst & oc20-s2ef & 0.149/0.128/0.128 & 0.125/0.125/0.125 & 0.083/0.083 & 0.006/0.006 & 0.699 & 0.535/0.545/0.545/0.548 \\
         & oc20-is2rs & - & - & - & - & - & - \\
         & oc20-is2re & \multicolumn{1}{l}{0.143/0.163/0.126} & \multicolumn{1}{l|}{0.132/0.156/0.123} & \multicolumn{1}{l}{0.094/0.069} & \multicolumn{1}{l|}{0.055/0.024} & 0.7393 & \multicolumn{1}{l|}{0.518/0.541/0.541/0.543} \\
         & oc22-s2ef & 0.272/0.238/0.235 & 0.156/0.139/0.148 & 0.256/0.254 & 0.477/0.476 & 0.6716 & 0.515/0.532/0.531/0.535 \\
         & oc22-is2res & - & - & - & - & - & - \\
         & oc22-is2re & \multicolumn{1}{l}{0.260/0.352/0.257} & \multicolumn{1}{l|}{0.242/0.328/0.137} & \multicolumn{1}{l}{0.315/0.250} & \multicolumn{1}{l|}{0.593/0.556} & 0.6720 & \multicolumn{1}{l|}{0.516/0.552/0.548/0.536} \\
        Polianna & article-level & 0.967/0.974/0.920 & 0.989/0.990/0.987 & 0.986/0.974 & 0.181/0.179 & 0.286 & 0.801/0.708/0.682/0.725 \\
         & text-level & - & - & - & - & - & -
        \end{tabular}
    }
    \caption{\label{table:analysis_scores} An analysis of dataset scores. Sample imbalance (SImb-score):= estimates selection biases; Spatio-temporal out-of-distribution (STood-score):= estimates Epistemic uncertainty; Input-output sensitivity (IO-score):= estimates Aleatoric uncertainty; Subgroups and edge cases (Outlier-score):= estimates evaluation biases.}
\end{table}

\subsubsection*{Random Forest performance benchmarks}

Table \ref{table:results_benchmark} contains the RF benchmark results computed for a standard tree maximum depth of 20 and a standard tree count of 128 for the ensembles. The sample ratio for each sub-task has been adjusted to moderate the computational (time and memory) complexity for training the benchmarks.

Except for ClimART and Open Catalyst, the RF results have an accuracy of less then 50\%. R2-scores have negative values for two sub-task datasets in Building Electricity and Uber Movement, meaning that the model performs worse than predicting the average of all labels. We obverse high variability in computational time for training our RF model per 1,000 data points, as this depends on the dimension of labels and features. However, we also observe small fluctuations in computational time for solving different tasks with the same feature and label dimensions. The electric power consumption is 3-4.3 time higher at the RAM than at the CPU. The performance of our RF model further has no clear correlation to any of the dataset scores we compute.

\begin{table}[ht]
    \scalebox{0.85}{
        \begin{tabular}{|l|l|r|c|c|c|l|}
        Task & dataset & \multicolumn{1}{c|}{\begin{tabular}[c]{@{}c@{}}sample\\ ratio\end{tabular}} & \begin{tabular}[c]{@{}c@{}}max depth \\ RF trees\end{tabular} & \begin{tabular}[c]{@{}c@{}}R2-score/\\ accuracy\end{tabular} & \begin{tabular}[c]{@{}c@{}}energy\\ CPU/RAM\\ (Wh)\end{tabular} & \multicolumn{1}{c|}{\begin{tabular}[c]{@{}c@{}}time/\\ time per 1,000\\ data (s)\end{tabular}} \\ \hline
        Building Electricity & buildings-92 & 1 & 20 & -1 & 544.4/1,817.7 & 8,565 (4.8) \\
         & buildings-451 & 0.1 & 20 & 0.348 & 315.9/1,054.6 & 5,022 (5.7) \\
        Wind Farm & days-177 & 0.1 & 20 & 0.417 & 1,087.3/4,297.9 & 20,466 (166.3) \\
         & days-245 & 0.1 & 20 & 0.322 & 2,169.9/7,244.6 & 34,498 (392.6) \\
        Uber Movement & cities-10 & 1 & 20 & -0.324 & 2,428.3/8,107.3 & 38,606 (0.17) \\
         & cities-20 & 0.01 & 16 (64 trees) & 0.192 & 90.5/361.1 & 1,662 (0.14) \\
         & cities-43 & 0.001 & 16 (32 trees) & 0.264 & 116.7/504.8 & 2,319 (0.94) \\
        ClimART & pristine & 0.1 & 20 & 0.925 & 1,387.9/4,633.9 & 22,066 (37.8) \\
         & clear-sky & 0.05 & 20 & 0.915 & 972.9/3,248.1 & 15,467 (52.7) \\
        Open Catalyst & oc20-s2ef & 1 & 20 & 0.25 & 1,094/4,380 & 20,863 (0.15) \\
         & oc20-is2re & 1 & 20 & 0.599 & 1.0/3.4 & 16.7 (0.03) \\
         & oc20-is2rs & - & - & - & - & - \\
         & oc22-s2ef & 1 & 20 & 0.281 & 31/114 & 541 (0.06) \\
         & oc22-is2re & 1 & 20 & 0.548 & 0.1/0.3 & 1.76 (0.03) \\
         & oc22-is2rs & - & - & - & - & - \\
        Polianna & article-level & 1 & 20 & 0.216 & 0.02/0.08 & 0.449 (1) \\
         & text-level & - & - & - & - & -
        \end{tabular}
    }
    \caption{Dataset benchmark evaluation results. Sample ratio indicates the percentage of samples used to train the RF model; the standard number of trees is 128. Time contains in the seconds per 1,000 data points in parentheses.}
    \label{table:results_benchmark}
\end{table}

\section*{Discussion}

We present ETT-17, the first collection of ML tasks and datasets for tackling climate change by enhancing solutions to the renewable energy transition, consistent with FAIR dataset guidelines \cite{Wilkinson2016}. We provide all tasks and datasets in a unified data format, and analyse similarities and differences between the collected tasks, with the goal of facilitating the development of over-parameterized and multi-tasking AI solutions in this domain. Our analysis reveals a number of important properties:

We find that large variations in the available number of data points exist between different tasks and application domains. This suggests that large multi-tasking and transfer learning models can be highly important for solving real-world problems with small data availability, such as for climate and energy policy text analyses, by leveraging inductive biases for \emph{zero-}/\emph{few-shot generalization} learned from training on related tasks with higher data availability. This is something that \emph{large language models} utilize for solving tasks with an unpractical small number of specific training data \cite{Brown2020}.

We further find that designing over-parameterized ML models that achieve \emph{interpolation} and are further able to perform robustly (\emph{universal law of robustness}) must have an unpractical large number of parameters in the order of several billions to trillions for solving most tasks we collect. This suggests that effective AI solutions in these domains may have to leverage active ML techniques to reduce the number of training data to a highly informative subset \cite{Aryandoust2022} and ML models that predict only one label at a time like \emph{transformers} and variations of \emph{graph neural networks} to reduce label dimensions \cite{Vaswani2017}, such that reaching \emph{interpolation} and satisfying the \emph{universal law of robustness} is feasible with a significantly smaller and more practical number of parameters. Reed et al. for example train their generalist reinforcement learning transformer agent on a filtered set of episodes with returns that are at least 80\% of the expert return for a task, which utilizes both these ideas \cite{Reed2022}.  

Our estimates of selection biases (SImb-score), Epistemic uncertainty (STood-score), Aleatoric uncertainty (IO-score), and evaluation biases (Outlier-score) associated with each prediction task again reveal large variations between tasks of different application domains, but show high similarity among tasks of the same domain. This suggests that designing a large, one-fits-all, multi-tasking ML model that performs better than several custom ML models for each specific task can be very challenging. Reed et al. for example observe how training a large multi-tasking specialist model on single application domains, without any fine-tuning on single tasks, can outperform much larger generalist models trained for solving tasks from multiple application domains \cite{Reed2022}.

The RF benchmark results we create are not satisfactory for all tasks. One reason for this is the limited modeling power of our RF models. Another reason is that our RF models are not able to make a distinction between regular and space/time-variant features, therefore not leveraging the properties of the data representation we use. For this reason, more specialized neural network based models like Graph Neural Networks and Transformers can be expected to be more accurate. 

Overall, our results are consistent with those of previous studies, which investigate multi-task and transfer learning models for solving diverse problem settings in other domains \cite{Bommasani2022, Furuta2022, Reed2022}. We prepare all datasets with spatio-temporal ood data for validation and testing, consistent with previous work by \cite{Koh2021}. For the design of multi-task and transfer learning models, all tasks from an entire application domains can be held out and be seen as ood for evaluating models for zero-shot and few-shot generalization performance \cite{Reed2022}. For the general reliability and acceptability of our study, future work must collect a larger variety of tasks and datasets, and explore applications where data is missing and multi-tasking ML models can be important. We utilize spatial and temporal data components to put all tasks and datasets into a single, unified and canonical data representation, despite their different data modalities and objectives. Future work may enhance the unified data representation we propose with more informative, universal graph representations by exploring the addition of adjacency between spatial and temporal coordinates we currently use.

\section*{Methods}

\subsubsection*{Unified spatio-temporal data representation}

A common property among all tasks is that they can be expressed with data components in (virtual) time and space. We leverage this property to formulate a unified spatio-temporal data representation.

\begin{figure}[h]
    \centering
    \includegraphics[height = 10.3cm]{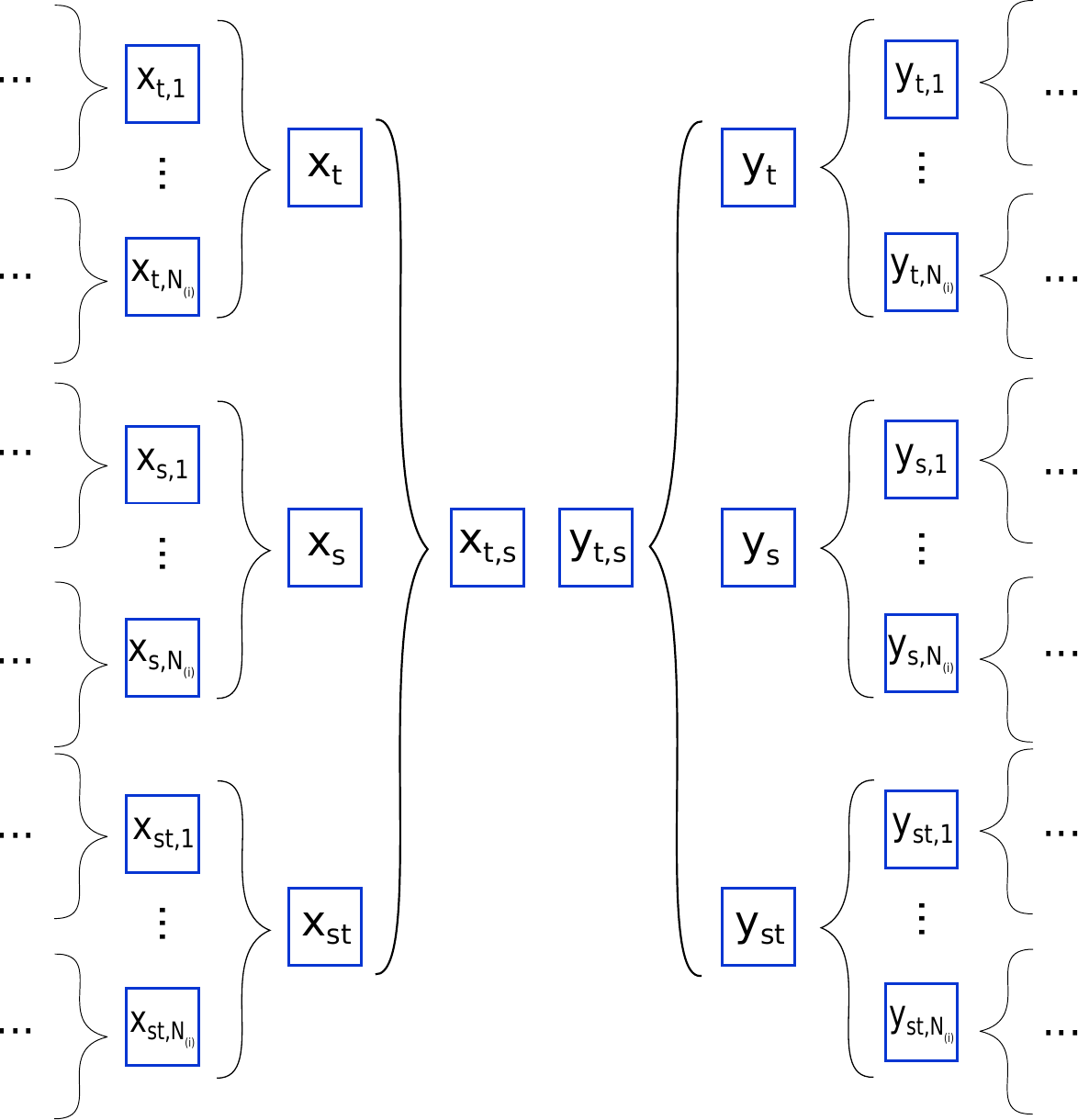}
    \caption{An illustration of our unified spatio-temporal data representation.}
    \label{figure:unified_format}
\end{figure}

Let every event in the world and a corresponding, single data point that captures it occur at a real or virtual coordinate in time $\mathbf{t} \in \mathbb{R}^{D_t}$ and/or space $\mathbf{s} \in \mathbb{R}^{D_s}$. Let every data point $(\mathbf{x}_{t,s} (, \mathbf{y}_{t,s}))$ further consist of a feature vector $\mathbf{x}_{t,s} \in \mathbb{R}^{D_x}$ and an optional label vector $\mathbf{y}_{t,s} \in \mathbb{R}^{D_y}$. We can then express $\mathbf{x}_{t, s} = concat(\mathbf{x}_{t}, \mathbf{x}_{s}, \mathbf{x}_{st})$ as a series of time-variant features $\mathbf{x}_{t}$ that are constant across space and only change across time; space-variant features $\mathbf{x}_{s}$ that are constant across time and only change across space; and space-time-variant features $\mathbf{x}_{st}$ that change across both space and time, with at least one of these having to be available for solving a given task. Let every available feature component $(i) \in \{t, s, st\}$ further consist of $N_{(i)}$ different sub-features, and every sub-feature $(j) \in \{1, \dots,  N_{(i)}\}$ further consist of potentially $M_{(i), (j)}$ different sub-sub-features, and so on. We can then write that $\mathbf{x}_{(i)} = concat(\mathbf{x}_{(i), 1}, \dots, \mathbf{x}_{(i), N_{(i)}})$ and $\mathbf{x}_{(i), (j)} = concat(\mathbf{x}_{(i), (j), 1}, \dots, \mathbf{x}_{(i), (j), M_{(i), (j)}})$, where $\mathbf{x}_{(i), (j)} \in \mathbb{R}^{D_{(i), (j)}}$ and $\mathbf{x}_{(i), (j), (k)} \in \mathbb{R}^{D_{(i), (j), (k)}}$, with $(k) \in \{1, \dots,  M_{(i), (j)}\}$, For the above to be valid, it further has to hold that $N_{(i)}, M_{(i), (j)}, D_t, D_s, D_x, D_y, D_{(i), (j)}, D_{(i), (j), (k)} \in \mathbb{Z}^{+}$. Let $\mathbf{y}_{t, s}$ be written in the very same manner whenever labels are available for a given dataset and task. Figure \ref{figure:unified_format} provides a visual illustration of this unified data representation.

The concept introduced here provides a universal framework for representing all types of datasets. By assigning unique coordinates in time and space to data points, we can utilize positional arguments in dynamic spatio-temporal graphs that are efficiently processed by tools like graph neural networks (GNNs). Likewise, we can efficiently handle spatio-temporal sequences with powerful models such as transformers. This approach permits optional labeling of data, thereby facilitating the expression of data for both supervised learning tasks (like regressions, classifications, object detection, or semantic segmentation) and unsupervised learning tasks (like clustering or dimensionality reduction) using the same unified data representation. We can express datasets from a specific task or domain in this representation using a canonical definition of coordinates in time and space for that task or domain.

Assuming the availability of at least one feature component in space-time, we can process a broad spectrum of applications by considering their data sparsity in time and space. Recognizing distinct sub-features for each feature type further allows us to distinguish between different data modalities that describe each feature type. Doing the same for labels further facilitates the use of distinguished loss functions during training. For instance, timestamps are a prevalent time-variant feature, uniform across space. They can typically be expressed through integers that represent quarters of an hour, hours of the day, or quarters of a year. To capture additional data describing a specific point in space at a given time, such as an image, we can employ a separate sub-feature. This method simplifies the processing through various (neural network) model architectures.

\subsubsection*{Over-parameterized ML}

The dimensionality of a dataset, that is, the number of features, labels and data points available for training an ML model, influences the optimal parametrization of a model and its (out-of-distribution) generalization performance, also called \emph{true risk} or \emph{testing accuracy}. To enable an informed decision about the efficient parametrization and sizing of ML models that we want to design for solving our prediction tasks, and achieve desired properties like associative memory \cite{Radhakrishnan2020} and robustness \cite{Bubeck2023}, we explore the dimensionality of each dataset. In particular, we want to know how many data points each dataset has; how these are split among training, validation and testing sets; and how large the number of each time-, space- and space-time-variant feature and label component of a data point is when expressed through our unified data representation. This allows us to compute an interpolation threshold $ipt$ for neural networks, that is, the number of model parameters necessary for falling into the region of over-parameterized DL or neural network prediction models as

\begin{equation*}
    ipt =  s_{tr}  n  D_{y},
\end{equation*}
    
where $n$ is the number of data points for a given task and $s_{tr}$ is the share of data points available for training. For variable length labels, we use the dimension of the largest label as $ D_{y}$. It further allows us to compute a smooth function threshold $sft$, which is the number of model parameters necessary to learn smooth/robust functions as

\begin{equation*}
    sft = ipt  D_{x} = s_{tr}  n  D_{y}  D_{x},
\end{equation*}

where we similarly use the dimension of the largest feature as $D_{x}$ for variable length features.

\subsubsection*{Dataset scores}

The SImb-score is computed on the feature space only. A uniform distribution of features is perceived as perfectly balanced, yielding a SImb-score of 0. We compute this as the average of the Jensen-Shannon divergence (JSD) between the distribution of each feature and the uniform distribution, which is a measure of the similarity between two probability distributions $P$ and $Q$, defined as
\[
JSD(P || Q) = \frac{1}{2} \left( KL\left(P || \frac{P+Q}{2}\right) + KL\left(Q || \frac{P+Q}{2}\right) \right),
\]
where KL(P || Q) is the Kullback-Leibler (KL) divergence from distribution P to Q, defined as
\[
KL(P || Q) = \sum_{i} P(i) \log \left(\frac{P(i)}{Q(i)}\right).
\]

The JSD is symmetric, and ranges from 0 to 1, where 0 indicates that the two distributions are identical, and 1 indicates maximum dissimilarity between the distributions. These two properties make the JSD a suitable distance metric for probability distributions and the scores that we yield for our datasets here. 

The STood-score computes the average JSD between the distribution of features in our training data and our validation or testing data. The \emph{spatio-temporal} component of the score originates from creating validation and testing datasets by splitting off data points from our total datasets across real or virtual coordinates in time and space, leveraging our unified data representation, which are then excluded from the resulting training data.

The IO-score captures the mean incremental ratio between all feature-label pairs as
\[
\frac{1}{D_{x}D_{y}} \sum_k^{D_{x}}\sum_m^{D_{y}} \frac{1}{|\alpha_{k, m}|} \sum_i^{|\alpha_{k, m}|} \frac{2}{\pi} \arctan |\frac{l_{k, m, i+1} - l_{k, m, i}}{f_{k, m, i+1} - f_{k, m, i}}|,
\]
where $|\alpha_{k, m}|$ is the cardinality of the set $ \alpha_{k, m} = \{ \alpha_{k, m, i} \}$ of all finite and defined incremental ratios, and $f_{k, m, i}$ are sorted feature values with corresponding labels $l_{k, m, i}$. Taking the absolute value of the incremental ratio allows us to avoid that sequential ratios with opposite signs reduce each other out. Furthermore, applying an $\arctan$ to the absolute value divided by $\frac{\pi}{2}$ allows us to obtain scores between 0 and 1. The $\arctan$ gives us the angle induced by the absolute value of the incremental ratio. The IO-score is an approximation of partial derivatives. However, when we account for changes in a single feature, it is crucial to acknowledge that other features might also change, which introduces a limitation in accurately approximating the partial derivative. This is because we are unable to isolate the variation of a single feature and observe its effect on a specific label. 

\begin{figure}[h!]
    \centering
    \includegraphics[scale=0.5]{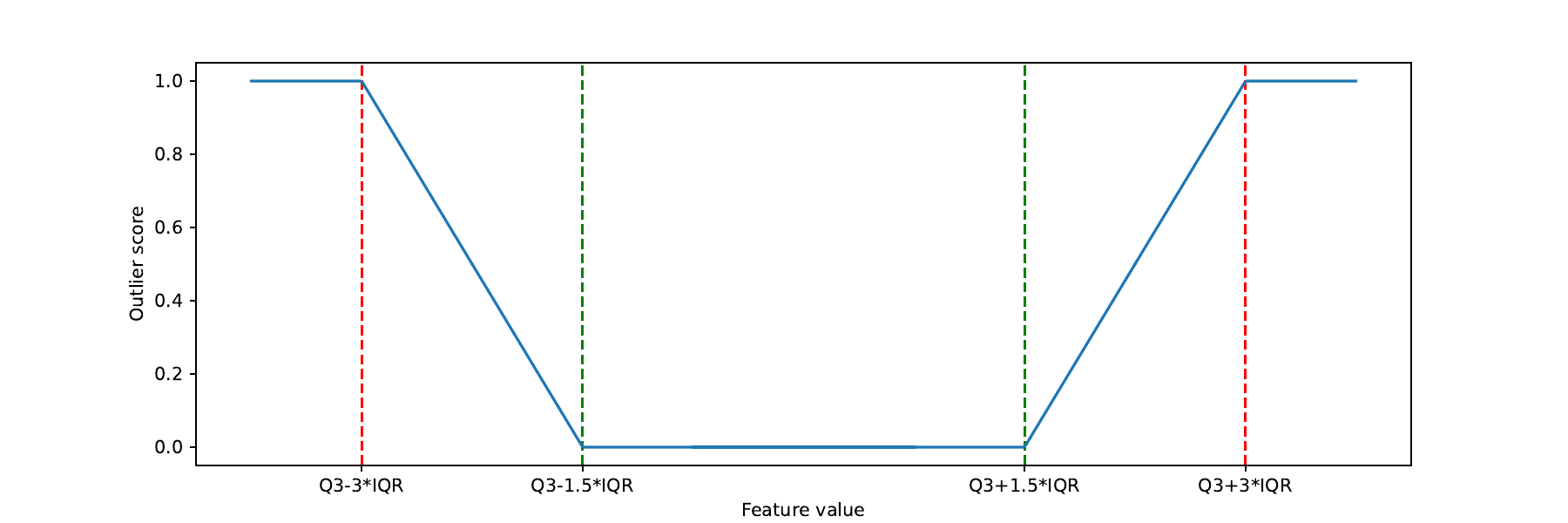}
    \caption{An illustration of the outlier scoring function used to compute the outlier score.}
    \label{figure:outlier_score}
\end{figure}

In order to address the potential impact of outliers on our SImb and STood scores, we applied the Tukey's fences method to identify and filter out these exceptional values. This approach utilizes the interquartile range (IQR) as a robust measure of statistical dispersion and employs it to establish two "fences" that separate the normal data from the outliers. We use the distribution of detected outliers to further define our Outlier-score. The key advantage of using Tukey's fences lies in its ability to handle outliers without imposing any specific assumptions about the underlying distribution of the data. By utilizing this technique, we ensure a more comprehensive analysis of the dataset while mitigating the influence of extreme values. Figure \ref{figure:outlier_score} illustrates the scoring function used to compute the Outlier-score.

The SImb-score is determined only on the input features of the data independently for each dataset split, ignoring the labels, using algorithm \ref{algo:simb_score}. The STood-score is computed similar to the SImb-score, relying on the approximation of feature distributions in the two considered splits (train-val or train-test) and then on the computation of the JS divergences between the distributions. We implement algorithm \ref{algo:stood_score} for both train-val splits and train-test splits. The computation of the IO-score proceeds as shown in algorithm \ref{algo:io_score}. Algorithm \ref{algo:outlier_score} further details the method we used to compute the Outlier-score based on the Tukey's fences methods and the scoring function shown in figure \ref{figure:outlier_score}.

\begin{algorithm}
    \caption{Compute SImb-score}
    \label{algo:simb_score}
    \begin{algorithmic}[1]
    
        \FOR{each feature in the input data}
            \STATE Compute the interquantile range (IQR) and the two Tukey's fences $Q1 - 1.5 * IQR$ and $Q3 + 1.5 * IQR$
            \STATE Remove outliers smaller than the first fence and greather than the second fence
            \STATE Compute a histogram with $10^4$ bins
            \STATE Normalize the histogram in such a way to obtain a distribution
            \STATE Compute the JS divergence between the computed distribution and a uniform distribution
            \STATE Store the computed score for iterated feature
        \ENDFOR
        
        \STATE Compute the SImb-score as the mean over all features' scores
        
        \RETURN SImb-score
    \end{algorithmic}
\end{algorithm}

\begin{algorithm}
    \caption{Compute STood-score}
    \label{algo:stood_score}
    \begin{algorithmic}[1]
        
        \FOR{each feature in the input data}
            \STATE Compute the interquantile range (IQR) and the two Tukey's fences $Q1 - 1.5 * IQR$ and $Q3 + 1.5 * IQR$
            \STATE Remove outliers smaller than the first fence and greather than the second fence
            \STATE Compute a histogram with $10^4$ bins for each of the two splits
            \STATE Normalize the two histograms in such a way to obtain two distributions
            \STATE Compute the JS divergences between the two computed distributions
            \STATE Store the computed score for the feature
        \ENDFOR
        
        \STATE Compute the STood-score as the mean over all features' scores
        
        \RETURN STood-score
    \end{algorithmic}
\end{algorithm}

\begin{algorithm}
    \caption{Compute IO-score}
    \label{algo:io_score}
    \begin{algorithmic}[1]
        \FOR{each pair (feature, label) in the data}
            \STATE Consider the sequence of sorted feature values $f_i$ and corresponding labels $l_i$
            \FOR{each pair of consecutive points}
                \STATE Compute the incremental ratio $\delta_i = \frac{l_{i+1} - l_i}{f_{i+1} - f_i}$ of the label with respect to the feature
                \IF{the incremental ratio is not infinite or undefined}
                    \STATE Compute $\alpha_i = \frac{2}{\pi} \arctan{|\delta_i|}$
                \ENDIF
            \ENDFOR
            \STATE Compute and store the mean value $\frac{1}{|\alpha|} \sum_i \alpha_i$, where $|\alpha|$ indicates the cardinality of the set $\alpha = \{\alpha_i\}$
        \ENDFOR
        
        \STATE Compute the IO-score as the mean over the feature-label pairs' scores
        
        \RETURN IO-score
    \end{algorithmic}
\end{algorithm}

\begin{algorithm}
    \caption{Compute Outlier-score}
    \label{algo:outlier_score}
    \begin{algorithmic}[1]
    
        \FOR{each feature in the input data}
            \STATE Compute the interquantile range (IQR), the two inner Tukey's fences $Q1 - 1.5 * IQR$ and $Q3 + 1.5 * IQR$, and the outer Tukey's fences $Q1 -3 * IQR$ and $Q3 + 3 * IQR$
            \STATE Compute a score for each data point using the piecewise scoring function shown in figure \ref{figure:outlier_score}
            \STATE Compute the feature mean score as the mean over all the data points' scores
        \ENDFOR
        
        \STATE Compute the outlier score as the mean over all features' scores
        
        \RETURN outlier score
    \end{algorithmic}
\end{algorithm}

\subsubsection*{Performance benchmarks}

For regression problems, we evaluate prediction performance using the \emph{R2-score} computed as

$$1 - \frac{\sum_i^N (y_i - \hat{y_i})^2}{\sum_i^N (y_i - \Bar{y})^2},$$
where $y_i$ are true labels, $\Bar{y}$ is their mean, $\hat{y_i}$ are predicted labels and $N$ is the number of evaluated samples. For classification problems, we evaluate prediction performance using accuracy computed as

$$\frac{1}{N} \sum_{i=1}^N \{y_i = \hat{y_i}\},$$
where $\{ \cdot \}$ is the indicator function for the predicted label $\hat{y_i}$ matching the true label $y_i$. 

The Polianna dataset has the problem that each sample is formed by a variable-length sequence of words, which a RF cannot directly learn. For this reason, we employ a common method based on word frequencies called the \emph{bag of words}. Each sample is converted into the count of occurrences of every word in the sentence. This leads to a fixed-length representation that we can feed into the RF model for classification. This approach has the advantage of being computationally lightweight, but does not retain the temporal information associated with the order of the words in the sentence. This means that we cannot employ it to solve the \emph{text-level} dataset task, where we need this temporal information to make predictions on sub-parts of an entire sample. 

An analogous problem appears in OpenCatalyst. Each molecule structure is formed by a variable-length list of atoms, where each atom is spatially set in space by 3-dimensional coordinates and each element has 8 + 22 additional features related to its physical and chemical properties. To solve this problem, the approach we adopt is similar to the one used in Polianna: the input to the RF model is formed by the counts of atoms in each molecule, juxtaposed to the mean and standard deviation of all the additional properties calculated for the atoms forming the molecule.
The dataset tasks \emph{-s2ef} and \emph{-is2re} consist of a 1-dimensional regression task that can be directly solved with a RF model, while the \emph{-is2rs} dataset tasks require the prediction of 3-dimensional positions of the relaxed structure of the molecule for each atom. This cannot be directly solved with our baseline model, since the model expects the outputs to have fixed length.

\subsubsection*{Dataset storage scheme}

In order to efficiently store data for our unified data representation, we need to map variables to separately stored files in two cases. In the first case, we map variables that come from a small set of diverse data points, but have a relatively large dimension. This allows us to store a single ID for a large number of data points and map these to the actual set of information during processing that is concisely stored separately. An example for such variables are the satellite image of a building in the \textit{Building Electricity} task, where only a set of 92-451 different buildings are given, while the number of data points ranges between 3-15 million. Replicating the image for each data point will result in unnecessarily large and redundant information storage/memory demand. 

In the second case, we map any variables that have an irregular length or structure to a separately stored file. An example are the texts that we process for the \textit{Polianna} task. This avoids the sparsity in data storage when using a tabular format. Such sparsity is particularly large where the difference between the average length and maximum length of a text is large. It allows us to provide the main datasets in a fixed-size, tabular .csv format and reduce memory demand to a minimum at a small cost of additional computation for the mapping during the usage of data. In applications where available memory is large and computation is the bottleneck, users can perform a (batch-wise) mapping before entering a training loop. For mapping target data that is regular, we store data in a tabular .csv format. For mapping target data that is irregular, we instead use a .json format.

\subsubsection*{\emph{Building Electricity}}

Aryandoust et al. predict the electric load profile of single buildings for a future time window of 24h in 15-min steps, given purely remotely sensed features consisting of the aerial image of a building and the meteorological conditions in the region of a building for a past time window of 24h in 1-h steps \cite{Aryandoust2022}. Among many different types of electric load forecasts that are useful for enhancing the energy transition, this falls into the category of short-term spatio-temporal load forecasts, and can be used for both planning and dispatch of renewable electricity systems. The primary goal is to assess the information content of candidate data points prior to physically measuring or streaming these, a method known as active learning, so as to reduce the need for smart meters and data queries to the most informative data only. 

The \emph{Building Electricity} dataset is different from other data in this domain, in that electric load is predicted from purely remotely sensed features, which makes ML solutions highly scalable. The authors publish their datasets under a MIT license (\url{https://opensource.org/licenses/MIT}) in a Harvard Dataverse repository (\url{https://doi.org/10.7910/DVN/3VYYET}).  The authors publish two sub-task datasets, one containing 92 buildings which we refer to as the \emph{buildings-92} dataset, and the other containing 451 buildings which we refer to as the \emph{buildings-451} dataset. 

Features and labels have equal structure for both sub-tasks. In order to anonymise their data, the authors provide only histograms of pixel values of aerial building images using 100 bins for each of the three RGB (red-green-blue) channels of an image. Features $\mathbf{x} = concat(\mathbf{x}_{t}, \mathbf{x}_{s}, \mathbf{x}_{st})  \in \mathbb{R}^{521}$ consist of a time-variant component $\mathbf{x}_{t} \in \mathbb{N}^{5}$ containing information on year, month, day, hour and quarter of hour (15-min steps), a space-variant component $\mathbf{x}_{s} \in \mathbb{N}^{300}$ containing the histogram of RGB (red-green-blue) channels of aerial images of a building with a bin size of 100 per channel, and a space-time-variant component $\mathbf{x}_{st} \in \mathbb{R}^{216}$ containing 9 meteorological condition time series of the past 24h in 1-h steps. Labels $\mathbf{y} = \mathbf{y}_{st} \in \mathbb{R}^{96}$ consist of the electric consumption profile of the given building at the given time for the next 24h in 15-min. 

Meteorological data are provided in time series of 15-min resolution so as match the electric load data resolution when splitting datasets. These are, however, replicated by a factor of 4 from an original resolution of 1-h. The meteorological values consist of air density in kg/m³, cloud cover, precipitation in mm/hour, ground level solar irradiance in W/m², top of atmosphere solar irradiance W/m², air temperature in °C, snowfall in mm/hour, snow mass in kg/m² and wind speed. We extract \emph{ood} data for testing and validation, by choosing all data points that belong to 20\% randomly sampled buildings, and have at least one of the five time stamp component that match 15\% randomly sampled points in time. Further detail on these datasets is given in the original study \cite{Aryandoust2022}.

\subsubsection*{\emph{Wind Farm}}

Zhou et al. present a prediction task, in which the active power generation of a wind farm has to be predicted for two days into the future, given two weeks of historic values of each turbine on the farm measured by a supervisory control and data acquisition (SCADA) tool at a given time and the relative geographic position of all turbines in the farm to each other \cite{Zhou2022}. Wind power has naturally high variability, and knowing the accurate generation of wind farms as well as their expected uncertainties is a great advantage for integrating high shares of these in our power systems at a higher pace, while planning and guaranteeing stable dispatch. In particular, more accurate predictions allow us to better balance out consumption and generation of electricity, where fluctuations in wind power require substitutions from other sources that might not be available at short notice, or where the distribution, sizing and control of substituting capacities needs to be planned and built out. 

The \emph{Wind Farm} dataset is particularly useful compared to other data in this domain, because the spatial distribution and internal state of each turbine on the wind farm is provided, and the number of turbines is comparably large. Another important  characteristic is, that measurement errors of the SCADA system are included, which requires solutions to be able to deal with sparsity in data. The authors publish SCADA data of a wind farm with 134 wind turbines for 245 and 177 days without a license on a public challenge website (\url{https://aistudio.baidu.com/aistudio/competition/detail/152/0/introduction}). The original application comes with no sub-tasks. Here, we split the task into data published for training, which we refer to as the \emph{days-245} sub-task and data that was used for hidden tests of the competition and revealed later, which we refer to as the \emph{days-177} sub-task. We introduce these sub-tasks because the authors use the same day counters in the original training and primarily hidden testing data, without revealing in what temporal relationship data points stand to each other across the two datasets. We split the continuous time series of measured data of a single turbine from the SCADA system into data points consisting of a past time series of all values (features) and the future time series of active power generation (labels). 

Features $\mathbf{x} = concat(\mathbf{x}_{t}, \mathbf{x}_s, \mathbf{x}_{st}) = concat(concat(\mathbf{x}_{t, 1}, \mathbf{x}_{t, 2}), \mathbf{x}_s, \mathbf{x}_{st}) \in \mathbb{R}^{4610}$ consist of a first time-variant component $\mathbf{x}_{t, 1} \in \mathbb{N}^{864}$ containing information about the day, hour of day and 10-min for each historic value ($288 * 3 = 864$), a second time-variant component $\mathbf{x}_{t, 2} \in \mathbb{N}^{864}$ containing the same information, but for each requested label of the future time window, together forming $\mathbf{x}_{t} = concat(\mathbf{x}_{t, 1}, \mathbf{x}_{t, 2}) \in \mathbb{R}^{1728}$; a space-variant component $\mathbf{x}_{s} \in \mathbb{R}^2$ containing the relative 2D position of a wind turbine in meters in Cartesian coordinates; and a space-time-variant component $\mathbf{x}_{st} \in \mathbb{R}^{2880}$ containing a past time window series of ten dynamic values measured by the SCADA system in each historic time stamp ($288 * 10 = 2880$). Labels $\mathbf{y} = \mathbf{y}_{st} \in \mathbb{R}^{288}$ consist of only the active power generation of the turbine in kW for the next 48h in 10-min steps. 

The original challenge requires a prediction window of 48 hours in 10-min steps, that is, 288 values for each turbine, and provides two weeks of historic values for this during hidden tests. We reduce the historic window to 288 values for each data point, so as to increase memory efficiency and encourage the design of solutions that require less context in a single data point for making predictions. The original data further includes measurement errors, so as to better represent real-world obstacles of the task, but evaluates these with a zero loss during testing. Here, we remove these data points, which makes the datasets sparse and instead include the time stamp of each available historic value and of each requested label prediction in a single data point, such that complete information on the resulting sparsity in maintained. We remove data by the following conditions, as presented by the original study \cite{Zhou2022}:

\begin{itemize}
    \item for active power $\leq$ 0 and wind speed $\geq$ 2.5: the actual active power is unknown.
    \item for pitch angles of either blade $>$ 89°: the actual active power is unknown.
    \item for nacelle directions $<$ -720° or $>$ 720°: anomaly of active power is given.
    \item for wind directions $<$ -180° or $>$ 180°: anomaly of active power is given.
\end{itemize}

Values measured by the SCADA system include wind speed recorded by the anemometer in m/s, angle between the wind direction and the position of turbine nacelle in °, temperature of the surrounding environment in °C, temperature inside the turbine nacelle in °C, the yaw angle of the nacelle in °, the pitch angles of all three blades in °, the reactive power generation in kW and the active power generation of the turbine in kW. We extract \emph{ood} data for testing and validation by choosing all data points that belong to 20\% of randomly sampled turbines, and have at least one of the 3 time stamp components that match 15\% randomly sampled points in time in blocks of at least 2 consecutive weeks.

\subsubsection*{\emph{Uber Movement}}

Aryandoust et al. use travel time changes between different zones of a city to infer probabilities of driving and choosing a destination, and sample spatio-temporal car parking maps on the scale of entire cities with these \cite{Aryandoust2019}. Their car parking maps are primarily used for planning the distribution and sizing of charging stations in renewable power systems when electrifying mobility and utilizing storage capacities for vehicle-to-grid electricity system services. We can predict origin-destination matrices of travel time between any two city zones at a given time, given the geographic information of city zones and information about circadian rhythms of a city. This allows us to expand the analysis of car parking maps to cities where no travel time data is available, or improve parking maps for cities where data is missing and travel time matrices are highly sparse. The Uber Movement project reports to state mean travel time data only where a statistically significant amount of trips is given. 

The \emph{Uber Movement} dataset is particularly interesting, because it is the most extensive public dataset for origin-destination matrices of mean travel time data for cities around the global. Missing data itself can therefore contain useful information on for example low traveling demand. Predicting missing travel time data hence bears the risks of worsening the performance of the downstream task for sampling parking maps when filling such informative data sparsity. The authors use travel time data published by the Uber Movement project (\url{https://movement.uber.com}) that are currently available under a CC BY-NC 3.0 US license (\url{https://creativecommons.org/licenses/by-nc/3.0/us/}) for 43 cities around the globe. The geographic information of city zones is further available under a custom license for each city. We divide all available data into three sub-tasks containing data from 10, 20 and 43 cities which we respectively refer to as the \emph{cities-10}, \emph{cities-20} and \emph{cities-43} datasets. Each consecutive sub-task contains all data from the previous one.

Features and labels have the same structure for both datasets. Features $\mathbf{x} = concat(\mathbf{x}_{t}, \mathbf{x}_{s}) = concat(\mathbf{x}_{t}, concat(\mathbf{x}_{s, 1}, \mathbf{x}_{s, 2})) \in \mathbb{R}^{11}$ consist of a time-variant component $\mathbf{x}_{t} \in \mathbb{N}^{4}$ with data on year, quarter of year, day type and hour of day; and a space-variant component $\mathbf{x}_{s} = concat(\mathbf{x}_{s, 1}, \mathbf{x}_{s, 2}) \in \mathbb{R}^{7}$ with data $\mathbf{x}_{s, 1} \in \mathbb{N}^{1}$ on city, and $\mathbf{x}_{s, 2} \in \mathbb{N}^{6}$ on origin-destination pairs of (x, y, z) coordinates. Labels $\mathbf{y} \in \mathbb{R}^{4}$ consist of the mean and standard deviation, as well as the geographic mean and geographic standard deviation of travel times between the given origin-destination pair of zones at the given point in time. 

City zones are described by geographic latitude and longitude coordinates of polygons, which may have largely different numbers of edges. We substitute this information with the centroid (\emph{lat\textsubscript{c}}, \emph{long\textsubscript{c}}) coordinates of city zone polygons (equation 1), which can be used for a regular representation of a polygon for the entire dataset. We use 

\begin{equation}
\begin{split}
    a &= \frac{1}{2} \sum_{i=0}^{N-1} (lat_{i} \; long_{i+1} - lat_{i+1} \; long_{i}) \\
    lat_c &= \frac{1}{6 a}  \sum_{i=0}^{N-1} (lat_{i} + lat_{i+1}) \; (lat_{i} \; long_{i+1} - lat_{i+1} \; long_{i}) \\
    long_c &= \frac{1}{6 a}  \sum_{i=0}^{N-1} (long_{i} + long_{i+1}) \; (lat_{i} \; long_{i+1} - lat_{i+1} \; long_{i}),
\end{split}
\end{equation}

where \emph{a} is the area of the polygon, \emph{N} is the number of edges of the polygon, and (\emph{lat\textsubscript{i}}, \emph{long\textsubscript{i}}) are the geographic coordinates of edge \emph{i}. We further transform the 2D geographic (\emph{lat}, \emph{long}) coordinates into a 3-dimensional unit sphere with (\emph{x}, \emph{y}, \emph{z}) coordinates (equation 2), using

\begin{equation}
\begin{split}
    x &= cos(lat) \; cos(long) \\
    y &= cos(lat) \; sin(long) \\ 
    z &= sin(lat).
\end{split}
\end{equation}

Without substituting $\mathbf{x}_{s, 2}$ with centroids, the geographic coordinates of each city zone polygon edge for any pair of origin-destination zone would lead to an irregular feature vector with dimensions between 30 and 53,646. Given the large number of data points available in this task, this is a required step for making data processing computationally feasible. We extract \emph{ood} data for testing and validation, by choosing all data points that belong to 20\% randomly sampled cities and city zones, and to 22\% of time stamp components. Detail on the down-stream applications that are relevant for the energy transition are provided in the original study \cite{Aryandoust2019}.

\subsubsection*{\emph{ClimART}}

Cachay et al. predict the atmospheric radiative transfer between different layers of the atmosphere using a simulation of different concentration of gases and aerosols at these with the Canadian Earth System Model \cite{Cachay2021}. Although the functional relationships used in climate and weather models are based on well-understood physics, values like the atmospheric radiative transfer are complex to calculate and create a computational bottleneck for modeling weather and climate at high resolutions in time and space. AI and ML surrogate models can emulate these relationships merely based on inputs and outputs and have the advantage of being significantly faster at inference time as compared to the original physics-based models. This allows us to model weather and climate at much higher resolutions in both time and space, and update our calculations more frequently as we measure new ground truth values over time. Improved climate projections are crucial for the planning process of renewable electricity systems. They allow us to better understand weather regimes such that we can for example reduce the demand for storage and generation capacities, achieve energy autarky in regions where it is physically or politically necessary, and guarantee the security of supply in time and space where climate and weather-dependent electricity generation introduces large fluctuations and uncertainty. 

The \emph{ClimART} dataset has more than 10 million data points, and is the most comprehensive dataset among all publicly available data in this field. Using such surrogate models, however, bears the risk of introducing unnoticed model biases. For example, AI and ML surrogate models with small bias errors might perform well for short-term weather predictions, but can have large accumulating errors for long-term climate projections. Similarly, AI and ML surrogates with random errors may perform well on long-term, low resolution projections of climate, but have significant errors for short-term, high-resolution weather forecasting. 

The authors provide data on the inputs, which is a vertical profile of atmospheric-surface properties, and outputs, which are shortwave and longwave (up- and down-welling) flux and heating rate profiles, of the Canadian Earth System Model (CanESM5). The data covers 40 years: 1850-52, 1979-1991, 1994-2014, and 2097-99. It is discretized into a geographic grid of 128 units of longitude and 64 units of latitude, giving (128 * 64 =) 8,192 points in space with 43 global snapshots (in 205h time steps) per year, and published under a CC BY 4.0 license (\url{https://creativecommons.org/licenses/by/4.0/}) on a self-hosted remote repository that can be accessed with instructions provided on Github (\url{https://github.com/RolnickLab/climart}). The authors publish two sub-tasks, which we refer to as the \emph{pristine-sky} dataset, where only the concentrations of gases are considered for predicting radiative transfer between different levels of the atmosphere, and the \emph{clear-sky} dataset, where also the concentration of aerosols is considered, that is, particles present in the air such as sulfur-containing compounds. 

Features and labels are available for 50 layers of the atmosphere and 49 levels between these layers, as well as globally on top of each profile. For every data point, that is, every 205h steps in time and 8'192 points in space, there exist $\mathbf{x} = concat(\mathbf{x}_{t}, \mathbf{x}_{s}, \mathbf{x}_{st}) \in \mathbb{R}^{970}$ features for the pristine-sky sub-task, $\mathbf{x} = concat(\mathbf{x}_{t}, \mathbf{x}_{s}, \mathbf{x}_{st}) \in \mathbb{R}^{2487}$ features for the clear-sky sub-task, and $\mathbf{y} = \mathbf{y}_{st} \in \mathbb{R}^{298}$ labels for both sub-tasks. Features consist of time-variant values $\mathbf{x}_{t} \in \mathbb{N}^2$ containing information about year and hour of year, space-variant features $\mathbf{x}_{s} \in \mathbb{R}^3$ containing information about 3-dimensional geographic \emph{(x, y, z)} coordinates on a unit sphere, and space-time-variant features $\mathbf{x}_{st} \in \mathbb{R}^{965}/\mathbb{R}^{2482}$ describing the physics and chemistry of the atmosphere and the Earth's surface. 

The original data comes with a number of caveats. Files that are provided by the authors of the original study for the years 1995 and 1851 are damaged at the time of writing this, which is why we exclude these here. A large number of data points further have outlying values for labels, in particular for solar and thermal heating rate profiles. We identify these as values that are greater than $10^{18} $ or smaller than $-10^{18}$. We augment data points by adding missing time stamp information to each data point containing year and hour of year. The original project stores inputs into the same files for both sub-tasks to avoid storage redundancy. It further uses a \emph{.h5} hierarchical data format to efficiently store the multidimensional arrays. Here, we create redundancy by separating the inputs for each sub-task and store all data in a \emph{.csv} format, where each row corresponds to a single data point, so as to simplify access and reduce the need for additional data processing for solving each sub-task. Further detail on these datasets are provided in the original study \cite{Cachay2021}. 

We create \emph{ood} data for testing and validation in time according to the recommendations of the original study, excluding pre-industrial and future data from the years 1850-52 and 2097-99 to test performance for unprecedented GHG concentrations and Earth surfaces, as well as the eruption of the Mount Pinatubo volcano in 1991 to test performance for a sudden, unprecedented increase in atmospheric opacity due to high-altitude volcanic aerosol. In order to avoid data leakage of the volcanic eruption which had lasting effects for a couple of years after the event during training, the original study does not provide data for the two consecutive years following the eruption in 1991. We extract further \emph{ood} data for testing and validation by separating a random subset of 20\% of hours per year and 35\% of geographic coordinates.

\subsubsection*{\emph{Open Catalyst}}

The \emph{Open Catalyst} projects presents the task of predicting the relaxed structure and energy of catalyst-adsorbate combinations for hydrogen electrolyzers and fuel cells from their initial structure, with the goal of enabling the large-scale exploration of new catalysts through efficient computational surrogates of density functional theory (DFT) simulations \cite{Zitnick2020, Chanussot2021, Tran2022}. Only the most promising candidates predicted by computationally efficient surrogate models can then be verified through more accurate DFT simulations before further exploration into their real-world feasibility. Green hydrogen, that is, hydrogen produced from excess (intermittent) renewable energy like wind and solar, can become a suitable solution to seasonable storage of electricity and therefore enhance the global energy transition. Its disadvantage compared to alternative storage systems like batteries and pumped-storage hydroelectricity is its low energy efficiency, which is currently at 35\% for round-trip AC to AC and unlikely to exceed 50\% \cite{Zakeri2015}. Finding new effective and low-cost catalysts to replace currently used expensive noble metals like iridium and platinum in both electrolysis and fuel cells is expected to reduce the cost of hydrogen storage technology and accelerate its adoption.

The \emph{Open Catalyst} project provides the largest datasets for hydrogen catalyst discovery, containing over a million adsorbate-catalyst relaxations that are randomly chosen from a pool of candidates, and 200 intermediate structures in their trajectory from initial to relaxed structures. One risk that exists for the energy transition when enhancing solutions to this task, however, is the potential use and advertisement of green hydrogen for applications that are more sustainable when using renewable energy directly. These are for example, hydrogen-ready boilers which have about one-sixth of the energy efficiency of heat-pumps for heating, and hydrogen fuel-cell electric vehicles which have about one-third of the energy efficiency of battery electric vehicles. Another risk that exists when enhancing hydrogen fuel cell efficiency is the simultaneous promotion of grey and blue hydrogen, which account for nearly all hydrogen used today, that can have higher GHG emissions through substantial CO\textsubscript{2} and fugitive methane emissions than the fossil oil and natural gas they may replace.

The authors publish all datasets under a CC BY 4.0 licence (\url{https://creativecommons.org/licenses/by/4.0/legalcode}) with instructions for download given in a public Github repository (\url{https://github.com/Open-Catalyst-Project/ocp/blob/main/DATASET.md}). There are two sets of data published under the names \emph{Open Catalyst 2020} and \emph{Open Catalyst 2022}, where the focus of the latter datasets is on Oxygen Evolution Reaction catalyst discovery. Each of these have three sub-tasks, which we refer to with the prefixes \emph{oc20-} and \emph{oc22-}. In the first sub-tasks which we refer to as \emph{oc20-s2ef} and \emph{oc22-s2ef}, the per atom forces and overall system energy has to be predicted from a 3-dimensional structure of adsorbate-catalyst atoms; this replaces computationally intensive DFT simulations, and is used in combination with optimization techniques for iteratively calculating the structures and energy throughout an entire relaxation process. In the second sub-tasks which we refer to as \emph{oc20-is2rs} and \emph{oc22-is2rs}, the relaxed structure has to be predicted from an initial structure of a system; this can be done directly, or in a diffusion process where intermediate structures of the entire trajectory are also considered. Solving this task allows us to then calculate also the relaxed energy using the predictors from the first sub-task (s2ef), or a single step of DFT calculation. In the third sub-tasks which we refer to as \emph{oc20-is2re} and \emph{oc22-is2re}, only the relaxed structure's energy has to be predicted from an initial structure of a system. 

Inputs consist of 3-dimensional atom positions and a set of per atom information, while outputs consist of 3-dimensional atom positions and forces, as well as a system energy. The resulting relaxed system energies are either directly or indirectly computed. They are used to compute the change in energy when the adsorbate comes in contact with the catalyst (adsorption energy). The adsorption energy is then used to predict the activation energies and eventually determine the desired rate of reactions for a given catalyst. 

The dimensions of features and labels are variable and depend on the number of atoms in a system of adsorbates and catalysts. Let therefore $A_{(i), j}$ be the number of atoms in structure $j$, $A^{min}_{(i)}$ be the minimum, and $A^{max}_{(i)}$ be the maximum number of atoms for sub-task $(i) \in$ \{-s2ef, -is2rs, -is2re\}. We can then write that features $\mathbf{x} = concat(\mathbf{x}_{s}, \mathbf{x}_{st}) = concat(concat(\mathbf{x}_{s, 1}, \mathbf{x}_{s, 2}, \mathbf{x}_{s, 3}), \mathbf{x}_{st}) \in (\mathbb{R}^{35  A^{min}_{(i)}} - \mathbb{R}^{35 A^{max}_{(i)}})$ consist of a space-variant component $\mathbf{x}_{s} = concat(\mathbf{x}_{s, 1}, \mathbf{x}_{s, 2}, \mathbf{x}_{s, 3}) \in \mathbb{R}^{32 A_{(i), j}}$ and a space-time-variant component $\mathbf{x}_{st} \in \mathbb{R}^{3  A_{(i), j}}$. Space-variant features contain numeric properties $\mathbf{x}_{s, 1} \in \mathbb{R}^{8  A_{(i), j}}$ on each atom; ordinally encoded information $\mathbf{x}_{s, 2} \in \mathbb{N}^{2  A_{(i), j}}$ about the standard state and group block of an atom; and one-hot encoded information $\mathbf{x}_{s, 3} \in \mathbb{N}^{22  A_{(i), j}}$ about oxidation states. Space-time-variant features consist of the 3-dimensional coordinates of each atom in Cartesian coordinates. The number of atoms $A_{(i), j}$ in all structures ranges between 7 and 235.

The structure of labels, in contrast, differs for each type of sub-task. For the $(i)=$ \{-s2ef\} tasks, labels $\mathbf{y} = concat(\mathbf{y}_{t}, \mathbf{y}_{st}) \in (\mathbb{R}^{1+3  A^{min}_{(i)}} - \mathbb{R}^{1+3 A^{max}_{(i)}})$ consist of a time-variant component $\mathbf{y}_{t} \in \mathbb{R}$ representing the system-level energy of the given structure, and a space-time-variant component $\mathbf{y}_{st} \in \mathbb{R}^{3 A_{(i), j}}$ representing 3-dimensional forces for every atom in Cartesian coordinates. In the $(i)=$ \{-is2rs,\} task, labels $\mathbf{y} = \mathbf{y}_{st} \in (\mathbb{R}^{3  A^{min}_{(i)}} - \mathbb{R}^{3 A^{max}_{(i)}})$ consist of a space-time-variant component representing the 3-dimensional location of every atom for the relaxed structure of the system in Cartesian coordinates. In the $(i)=$ \{-is2re\} task, labels $\mathbf{y} = \mathbf{y}_{t} \in \mathbb{R}$ consist of a single time-variant value describing the relaxed energy of the given catalys and adsorbate system.

We augment the datasets with information from the periodic table of elements that each atomic number can be mapped to. We use the periodic table provided on \url{https://pubchem.ncbi.nlm.nih.gov/periodic-table/}. We omit all trajectory data and save data for the two sub-tasks \emph{oc20-is2re} and \emph{oc20-is2rs}, as well as \emph{oc22-is2re} and \emph{oc22-is2rs} together. The numeric features of each atoms property are atomic mass, electronegativity, atomic radius, ionization energy, electron affinity, melting point, boiling point and density. The standard states of an atom can be 'gas', 'solid', 'liquid', 'expected to be a solid', or 'expected to be a gas'. The group block of an atom can be 'non-metal', 'noble gas', 'Alkali metal', 'Alkaline earth metal', 'metalloid', 'halogen', 'post-transition metal', 'transition metal', 'lanthanide' or 'actinide'. 

We ensure validation and testing data to be ood by separating these conditioned on spatial features. The number of atoms in each structure is between 7 and 235. We separate all data points that have a structure with the number of atoms falling within a range of [87, 155] ($=121 \pm 30\%$ of possible atoms).

\subsubsection*{\emph{Polianna}}

Sewerin et al. present a dataset of 412 labeled legislative articles from 18 EU climate change mitigation and renewable energy laws, with the primary goal of enhancing large and systematic assessments of climate and energy policies \cite{Sewerin2023}. Given the article text of a policy directive or regulation, the task is to annotate text spans in the article with a set of (i) policy instrument types, (ii) policy design characteristics, and (iii) technology and application specificity. These three policy design elements are increasingly identified to be important determinants of policy effectiveness by the public policy literature. Enhancing solutions to this task, which is also known as \emph{named entity recognition} in the field of natural language processing, allows us to develop new semi-automated analytical tools for assessing the effectiveness of climate and energy policy designs, and move away from resource-intensive hand-coding in this field. It will allow us to perform large-scale ex-post analyses of policies, and make more effective ex-ante recommendations for the energy transition.

\emph{Polianna} is the first machine-learnable dataset for identifying a causal link between the adoption of energy and climate policy design (policy output) and their impact on the renewable energy transition and mitigating climate change (policy outcome). The authors publish their dataset under an CC BY 4.0 license (\url{https://creativecommons.org/licenses/by/4.0/}) on Zenodo (\url{https://zenodo.org/record/7569275#.ZEWyB5FBziM}). We introduce two different sub-tasks that we refer to as the \textit{article-level} task and the \textit{text-level} task. In the \textit{article-level} task, we want to estimate the frequency of annotations for the overall text of an article, regardless of their precise positions with the text. This is a simplified version of the original and more fine-grained \textit{text-level} annotation task. In the \textit{text-level} task instead, we want to identify precise text spans within an article that belong to a particular annotation label.

Features $\mathbf{x} = concat(\mathbf{x}_t, \mathbf{x}_s, \mathbf{x}_{st}) \in \mathbb{R}^{17-4963}$ are the same for both sub-tasks. They consist of a time-variant component $\mathbf{x}_t  \in \mathbb{Z}^{3}$ that contains information about year, month and day of an article's publication; a component in virtual space $\mathbf{x}_s \in \mathbb{N}^{2}$ that contains information about the form of the article and the treaty associated with the article; and the main article text associated with these two other components, which we categorize here as a space-time-variant component $\mathbf{x}_{st} \in \mathbb{N}^{D_{(i)}}$, where $D_{(i)} \in \mathbb{Z}$ may correspond to both the number of characters or number of words in article (i). Let $D_{(i)}$ represent the number of words in article (i) here. Then, with the minimum of an article length being 12 words, and the maximum being 4,958 words, it holds that $\mathbf{x}_{st} \in (\mathbb{N}^{12} - \mathbb{N}^{4958})$ and $\mathbf{x} \in (\mathbb{R}^{17} - \mathbb{R}^{4963})$  We provide the latter in the form of a continuous string of characters, as well as a tokenized tuple including starting index, ending index and the text of each word in an article. 

Labels, however, are different for each sub-task. In the \textit{article-level} task, labels to be predicted are the count, or distribution, over all possible annotations in the entire article. We are given a total of 42 possible annotations plus the case where we are uncertain about the concrete categorization among these. This yields a $\mathbf{y} = \mathbf{y}_{st} \in \mathbb{R}^{43}$ for the \textit{article-level} task. In the \textit{text-level} task, we want to predict the precise location of each annotation. Theoretically, this will allow us to annotate up to every word in an article with some label, where the dimension of labels will be equal to three times the article size (start, end, tag), if every pair of neighbouring words have two distinct labels. The true minimum and maximum number of tags available in the Polianna dataset, however, are 1 and 736, yielding a label dimension of three times this, such that $\mathbf{y} = \mathbf{y}_{st} \in (\mathbb{R}^{3} - \mathbb{R}^{2208})$. 

The form of an article can be either a directive or regulation. The associated treaties can be either the \textit{'Treaty establishing the European Community (consolidated version 1992)'}, the \textit{'Treaty on the Functioning of the European Union (consolidated version 2008)'}, or \textit{'Treaty on the Functioning of the European Union (consolidated version 2012)'}. We create ood data for testing by separating all data points that are associated with one randomly sampled treaty (spatial feature) and 40\% of all available years (temporal feature).

\subsection*{Code and data availability}

We provide all datasets on a public repository on Harvard Dataverse, \url{https://dataverse.harvard.edu/dataverse/EnergyTransitionTasks}. Code we use to process raw data and analyze results is further available on Github, \url{https://github.com/ArsamAryandoust/EnergyTransitionTasks}. All results can be reproduced with Docker containers using the instruction we provide on Github. A Python package for downloading data is developed at \url{https://github.com/Selber-AI/selberai}.

\section*{Supplementary Information}

\subsection*{Supplementary Information 1 - Alternative unified data representations}
Unified data representations have mainly been developed to deal with multiple data modalities in multi-task, transfer learning models. \emph{Gato} \cite{Reed2022} for example uses natural language as a common grounding. All data modalities, including images, text, proprioception, joint torques, button presses and various other discrete and continuous actions or observations, are serialized into a flat sequence of tokens for this. First, a coding scheme is used to tokenize all information by encoding texts into ordinal integers representing words, images into raster-ordered patches, discrete values into sequences of row-major-ordered integers, and continues values into row-major-ordered floating points. Then, a canonical sequence ordering is used to serialize all tokens. All data is represented as a sequence of tokens, where a full sequence or single data point for training consists of a reinforcement learning episode of $T$ time steps. Each time step in turn is a concatenation of observations and actions taken based on these. A lexicographic order for token types (text-image-tensors) in a single observation then enables the use of multi-modal data as input and output to each task. Finally, all data is brought into a unified data representation by embedding all tokens into the same vector space and concatenating with two positional encodings that provide temporal and spatial information for every token to the model. To further disambiguate distinct tasks within a domain, which share identical data modalities, a prompt sequence is prepended to a subset of data points in each batch during training. The prompts that are used are for example "You are an image captioner.", or, "You are a chat bot." One can then view the autoregressive sequence of observations plus actions of past time steps as features and those of future time steps as labels of potentially variable lengths.

An advantage that our representation has compared to the above mentioned ones is that positional information, which we call \emph{coordinates}, are assigned to single data points in both time and space. This allows us to represent data points as a dynamic, spatio-temporal point process or event graph. In the existing sequential representations we describe above, only a temporal component is provided as a time step for single data points, and spatial information is only provided to assign a position within data points. Another advantage that our unified data representation has over existing ones is that sparse data can be processed without further issues: if a measurement/data point is missing at some point in time or space, we can index the next available information or even at any other arbitrary point, and readily provide an ML model with this information. Existing models, in contrast, provide only local observation position encodings to a model. This means that only data from consecutive local time steps can be identified by a model, always starting with an index 0 or 1, without being able to process missing/sparse data.

\subsection*{Supplementary information 2 - Score plots for all datasets}

\begin{figure}[ht]
    \centering
    \includegraphics[height = 16cm]{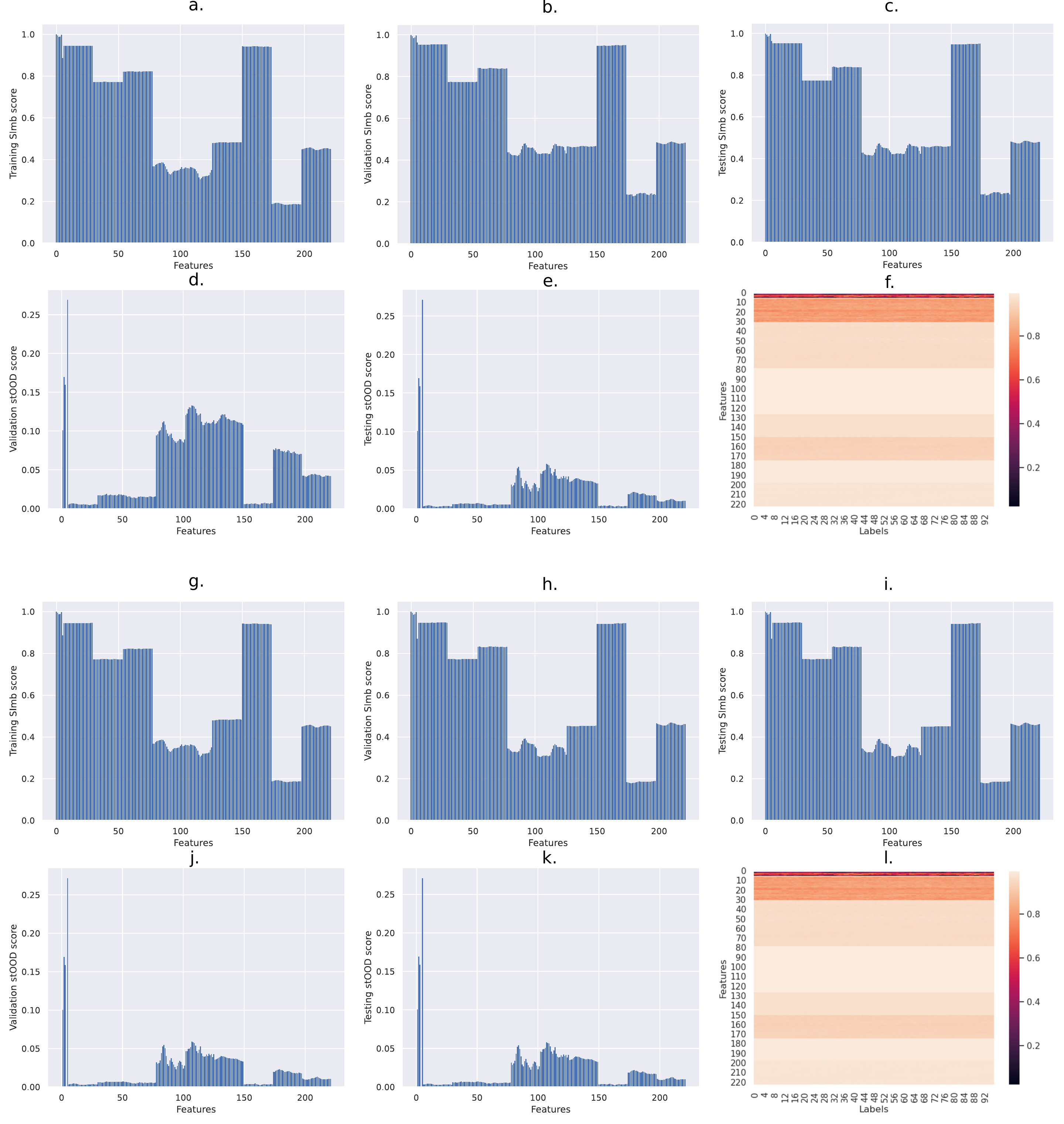}
    \caption{All dataset scores for Building Electricity task. Scores for sub-task dataset buildings-92 are represented in a.-f., while scores for sub-task dataset buildings-451 are represented in g.-l.}
    \label{figure:suppinf_be}
\end{figure}

\begin{figure}[ht]
    \centering
    \includegraphics[height = 16cm]{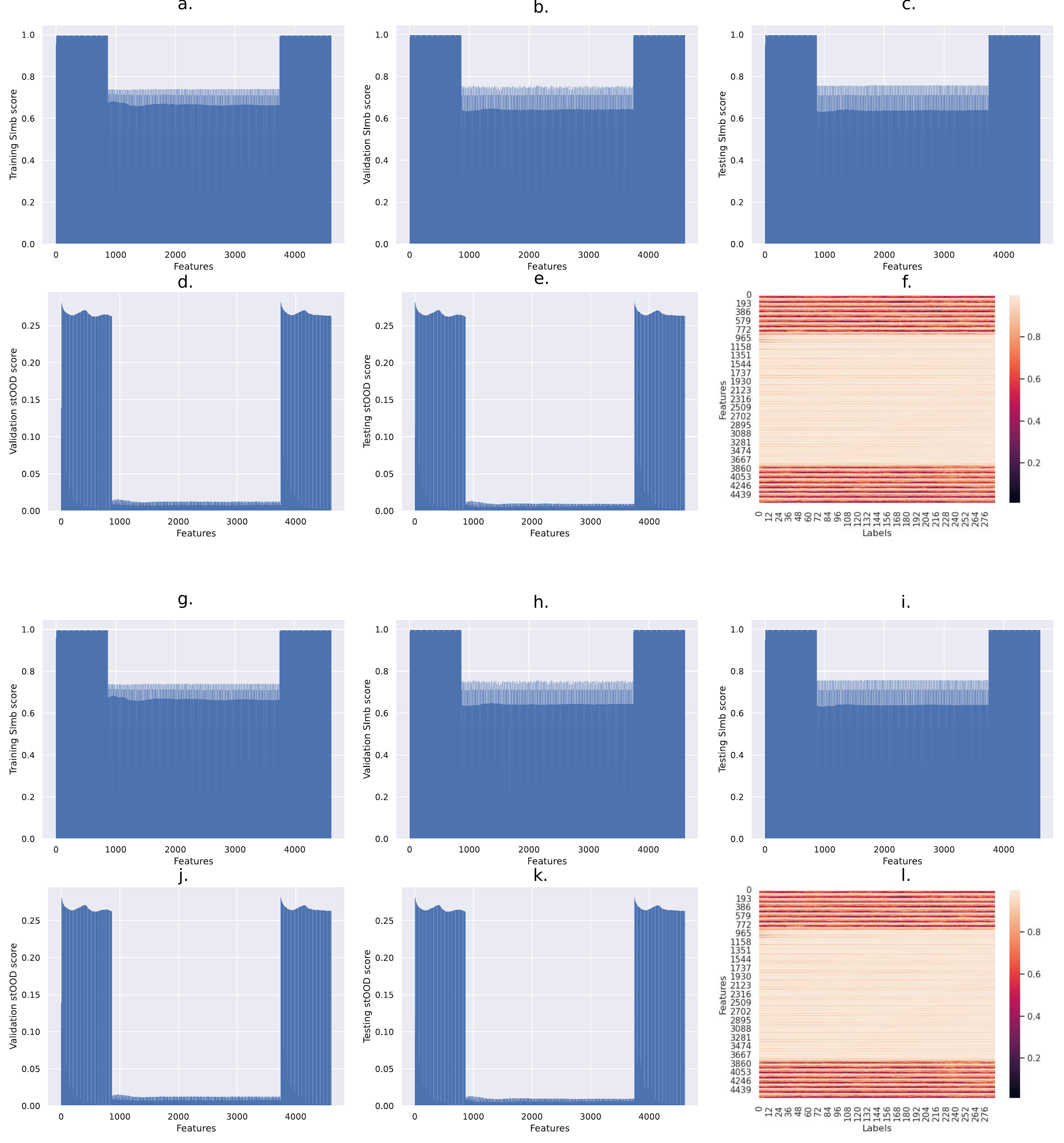}
    \caption{All dataset scores for Wind Farm task. Scores for sub-task dataset days-245 are represented in a.-f., while scores for sub-task dataset days-177 are represented in g.-l.}
    \label{figure:suppinf_wf}
\end{figure}

\begin{figure}[ht]
    \centering
    \includegraphics[height = 22cm]{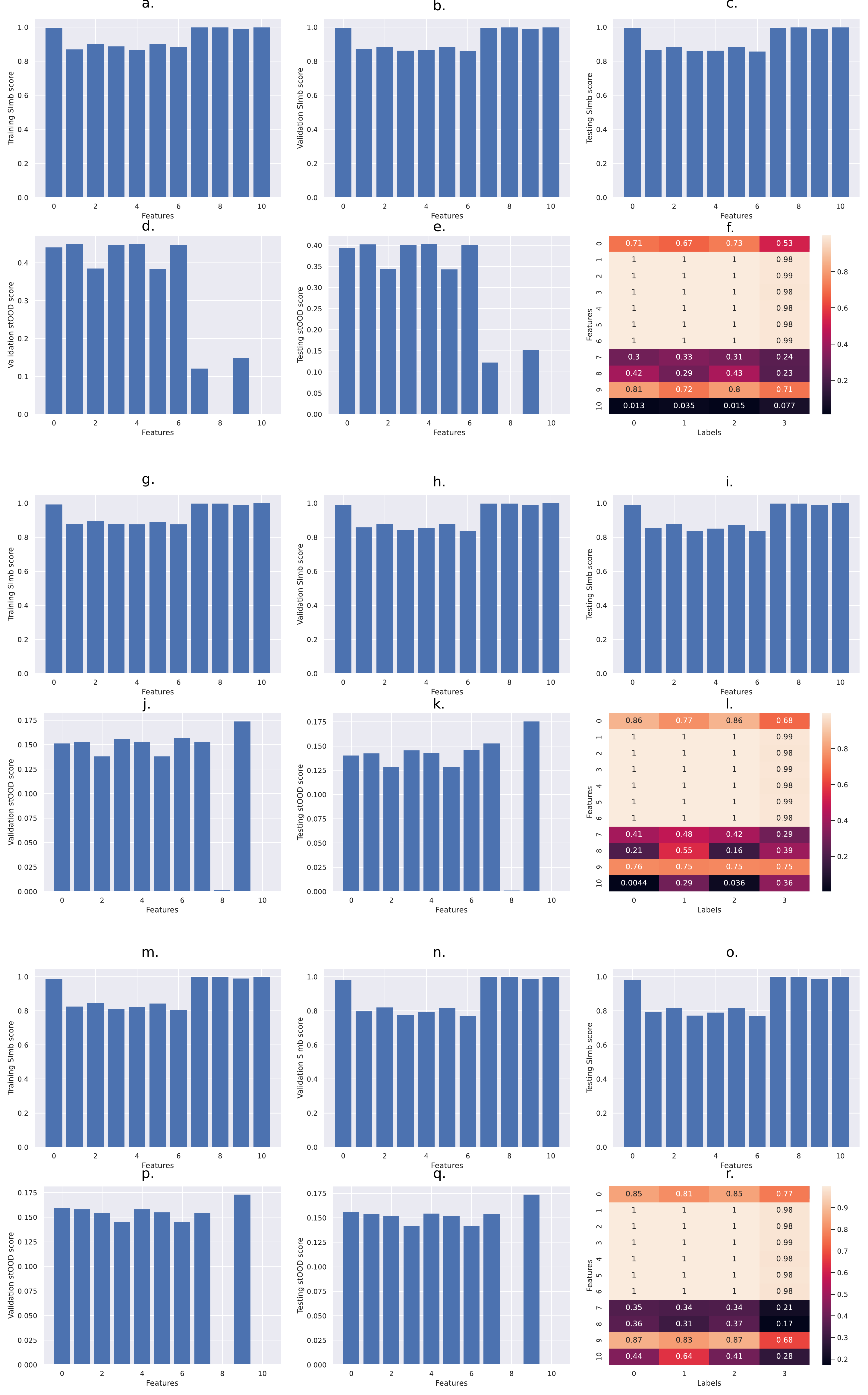}
    \caption{All dataset scores for Uber Movement task. Scores for sub-task dataset cities-10 are represented in a.-f., while scores for sub-task dataset cities-20 are represented in g.-l., and cities-43 are in m.-r.}
    \label{figure:suppinf_um}
\end{figure}

\begin{figure}[ht]
    \centering
    \includegraphics[height = 16cm]{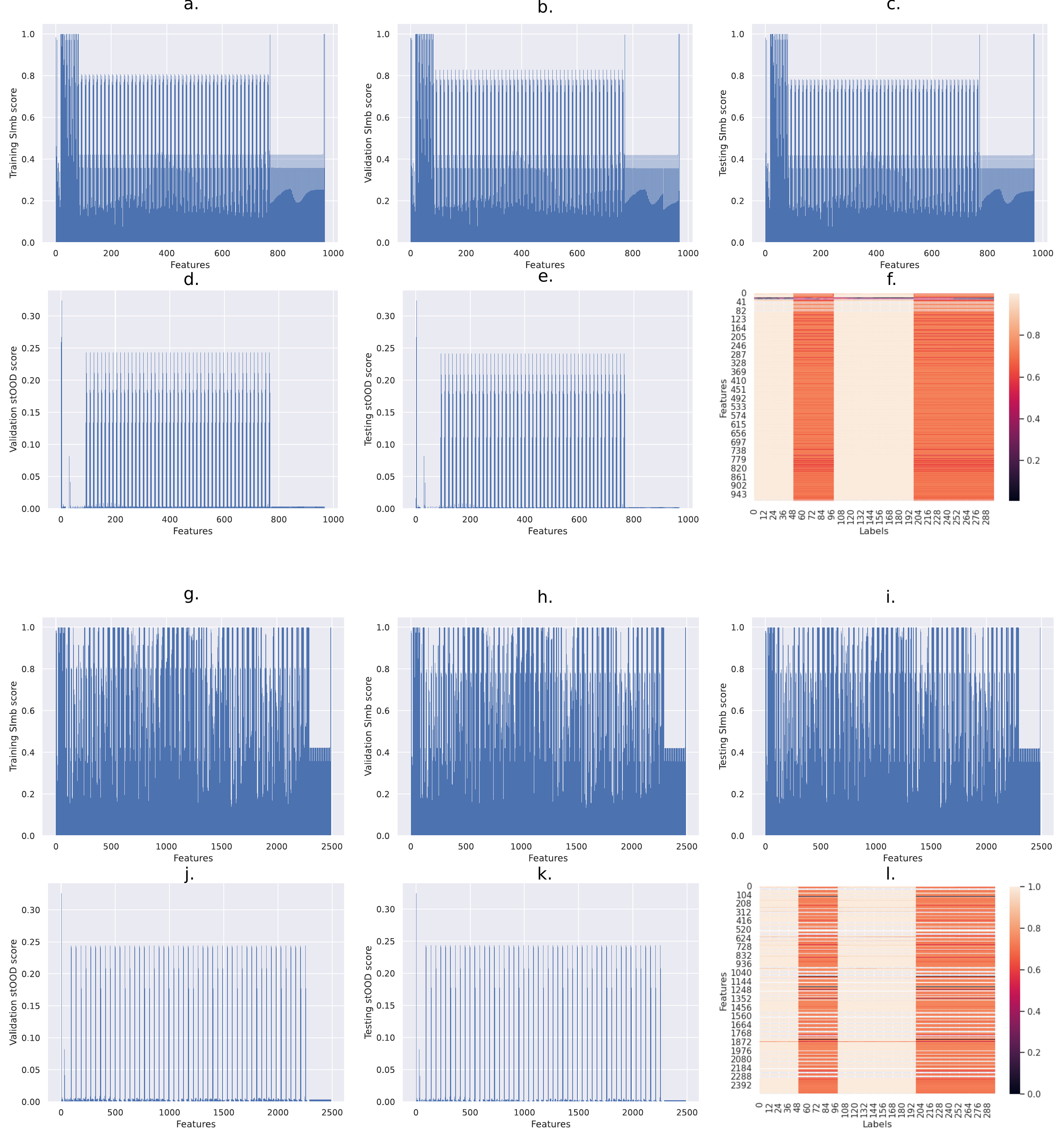}
    \caption{All dataset scores for ClimART task. Scores for sub-task dataset pristine-sky are represented in a.-f., while scores for sub-task dataset clear-sky are represented in g.-l.}
    \label{figure:suppinf_ca}
\end{figure}

\begin{figure}[ht]
    \centering
    \includegraphics[height = 16cm]{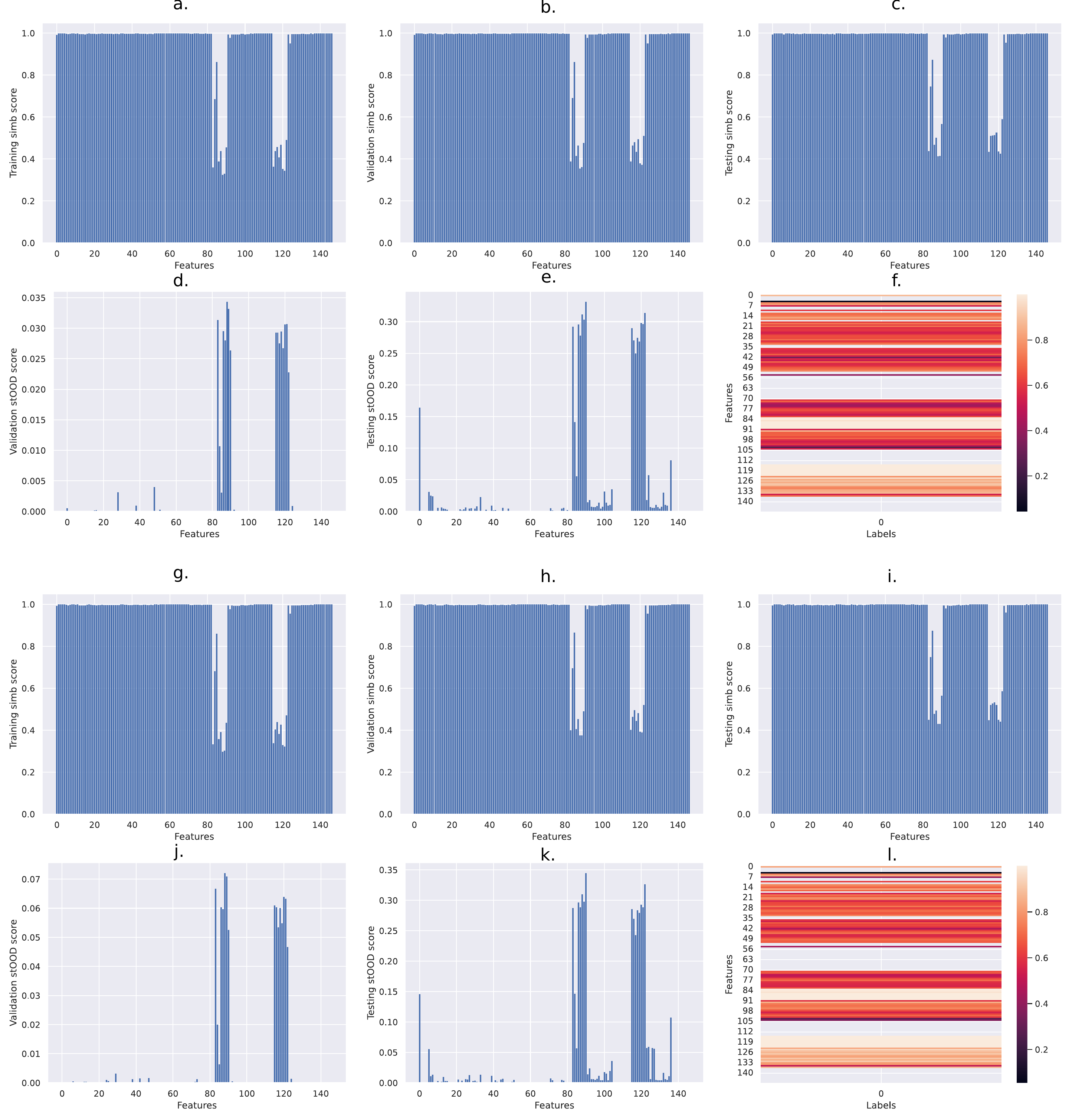}
    \caption{All dataset scores for Open Catalyst 2020 task. Scores for sub-task dataset oc20-s2ef are represented in a.-f., while scores for sub-task dataset oc20-is2re are represented in g.-l.}
    \label{figure:suppinf_oc20}
\end{figure}

\begin{figure}[ht]
    \centering
    \includegraphics[height = 16cm]{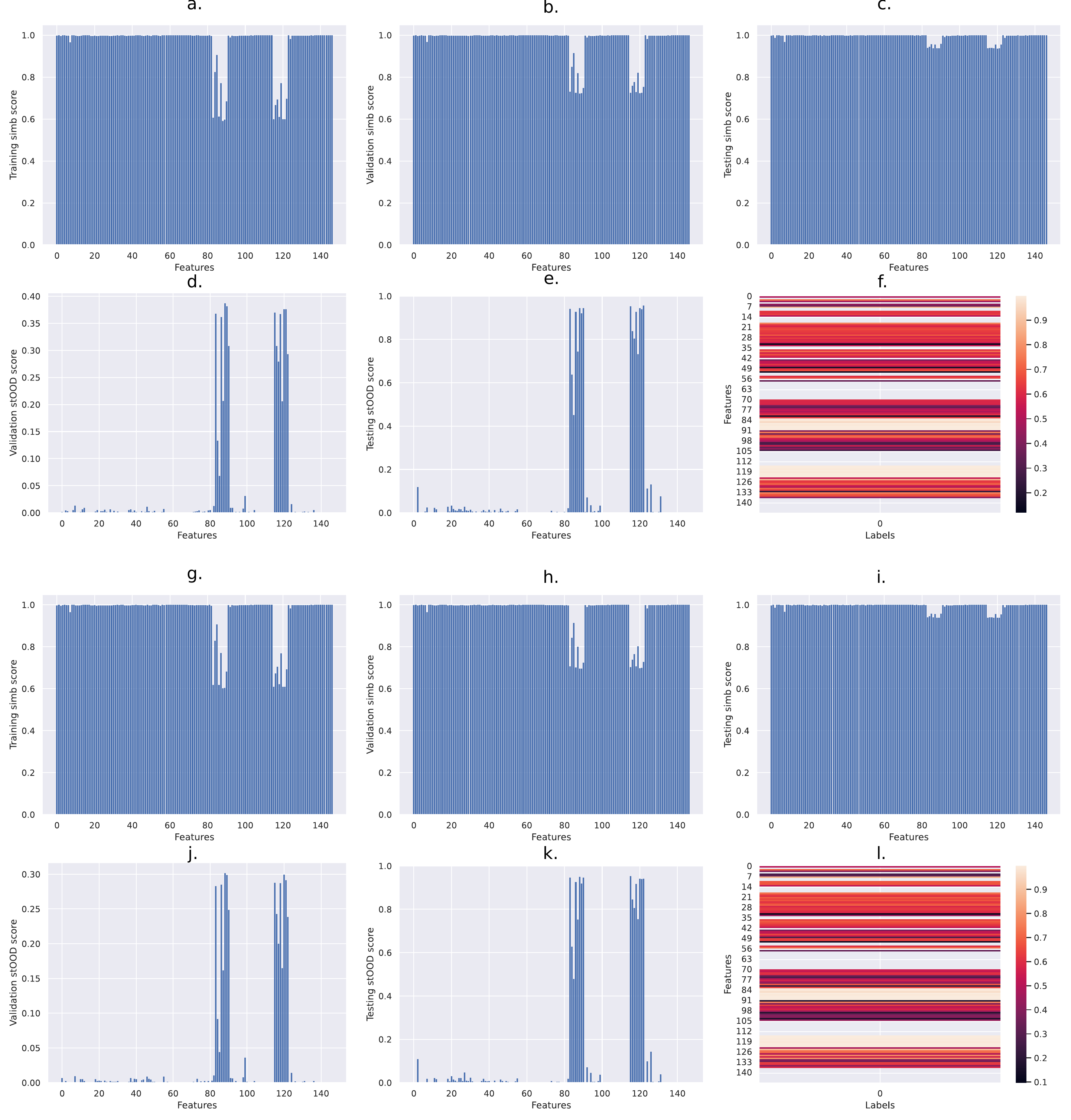}
    \caption{All dataset scores for Open Catalyst 2022 task. Scores for sub-task dataset oc22-s2ef are represented in a.-f., while scores for sub-task dataset oc22-is2re are represented in g.-l.}
    \label{figure:suppinf_oc22}
\end{figure}

\begin{figure}[ht]
    \centering
    \includegraphics[height = 8cm]{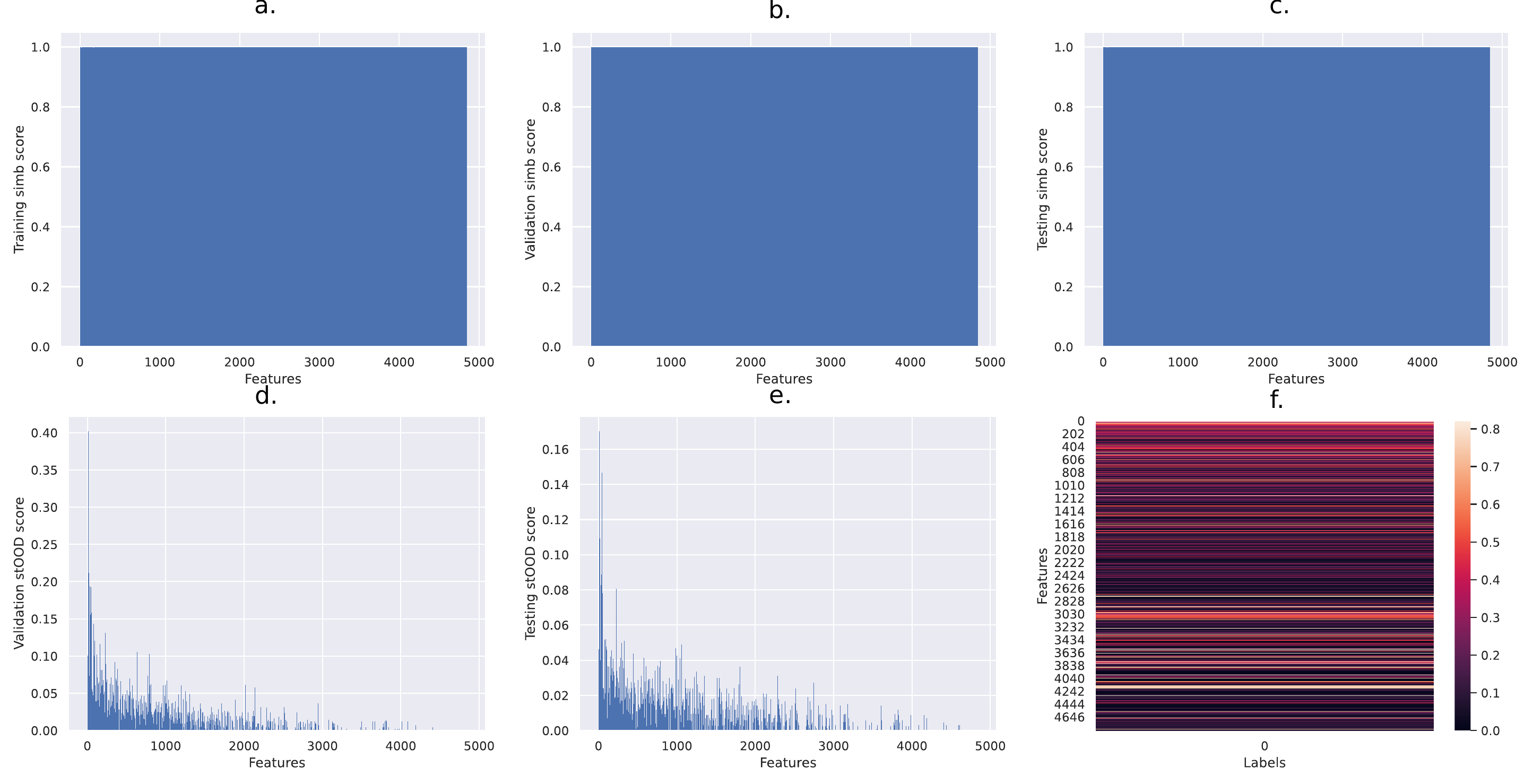}
    \caption{All dataset scores for Polianna task. Scores for sub-task dataset article-level are represented in a.-f..}
    \label{figure:suppinf_pa}
\end{figure}

Figures \ref{figure:suppinf_be}-\ref{figure:suppinf_pa} illustrate the results to all dataset scores for each task and sub-task dataset.

\end{document}